%% file: LiDARTag_main.tex
\documentclass[letterpaper, 10 pt, conference]{IEEEtran}  
\IEEEoverridecommandlockouts

\input{defines}

\title{LiDARTag: A Real-Time Fiducial Tag System for~Point~Clouds}
\author{Jiunn-Kai Huang, Shoutian Wang, Maani Ghaffari, and Jessy W. Grizzle
\thanks{J. Huang, S. Wang M. Ghaffari, and J. Grizzle, are with the Robotics
Institute, University of Michigan, Ann Arbor, MI 48109, USA. \texttt{\{bjhuang, shoutian, maanigj, grizzle\}@umich.edu}.} }

\begin{document}
\maketitle
\pagestyle{plain}
\begin{abstract} 
Image-based fiducial markers are useful in problems such as object tracking in cluttered or textureless environments, camera (and multi-sensor) calibration tasks, and vision-based simultaneous localization and
mapping (SLAM). The state-of-the-art fiducial marker detection algorithms rely on
the consistency of the ambient lighting. This paper introduces LiDARTag, a novel fiducial
tag design and detection algorithm suitable for light detection and ranging (LiDAR)
point clouds. The proposed method runs in real-time and can process data at 100
Hz, which is faster than the currently available LiDAR sensor frequencies. Because of the LiDAR sensors' nature, rapidly changing ambient lighting will not affect the detection of a LiDARTag; hence, the proposed fiducial marker can operate in a completely dark environment. In addition, the LiDARTag nicely complements and is compatible with existing visual fiducial markers, such as AprilTags, allowing for efficient multi-sensor fusion and calibration tasks.
We further propose a concept of minimizing a fitting error between a point cloud and
the marker's template to estimate the marker's pose. The proposed method achieves
millimeter error in translation and a few degrees in rotation.  Due to LiDAR returns' sparsity, the point cloud is lifted to a continuous function in a reproducing kernel Hilbert space where the inner product can be used to determine a marker's ID. The experimental results, verified by a motion
capture system, confirm that the proposed method can reliably provide a tag's pose and unique ID code. The rejection of false positives is validated on the Google
Cartographer indoor dataset and the Honda H3D outdoor dataset. All implementations
are coded in C++ and are available at:
\href{https://github.com/UMich-BipedLab/LiDARTag}{https://github.com/UMich-BipedLab/LiDARTag}.
\end{abstract}

\input{Introduction}

\input{Related}

\input{TagDesign}
\input{Detector}
\input{PoseEstimation}

\input{Decoding}
\input{Experiments}

\input{Conclusion}

\vspace{-2mm}
\section*{Acknowledgment}
Toyota Research Institute provided funds to support this work. Funding for J. Grizzle was in part provided by NSF Award No.~1808051. The first author thanks Wonhui Kim for useful conversations.


\bibliographystyle{bib/IEEEtran}
\bibliography{bib/strings-abrv,bib/ieee-abrv,bib/references}

\end{document}

%% file: defines.tex
\usepackage{bmpsize}
\usepackage{amsthm}
\usepackage{algorithm, algorithmicx, algpseudocode}
\usepackage{amsmath,amssymb,enumerate}
\usepackage{mathrsfs}
\usepackage{times}
\usepackage{multicol}
\usepackage{amsfonts}
\usepackage{textcomp}
\usepackage{balance}
\usepackage{multirow}
\usepackage{booktabs}
\usepackage{dblfloatfix} 
\usepackage{bbold}
\usepackage{makecell}
\usepackage{hhline} 
\usepackage{float} 

\usepackage{comment}
\usepackage{xparse}
\usepackage{lipsum}
\usepackage{soul} 
\usepackage{times}
\usepackage{cite}
\usepackage{balance}
\usepackage{dsfont} 
\usepackage{physics} 
\usepackage{setspace} 
\usepackage{accents} 

\usepackage[dvipdfmx]{graphicx} 
\usepackage[T1]{fontenc}
\usepackage[usenames,dvipsnames,svgnames,table, xcdraw]{xcolor}
\usepackage[caption=false,font=footnotesize]{subfig}
\usepackage[utf8]{inputenc}

\usepackage{hyperref}
\hypersetup{
  colorlinks=true,pdfpagemode=UseNone,citecolor=black,linkcolor=black,urlcolor=BrickRed
}

\usepackage{caption}
\graphicspath{
    {graphics/}
}

\newtheoremstyle{break}
  {\topsep}{\topsep}%
  {\itshape}{}%
  {\bfseries}{}%
  {\newline}{}%

\newtheorem{remark}{Remark}

\theoremstyle{break}

\definecolor{note}{rgb}{0.3,0.7,0.25}
\definecolor{rephase}{rgb}{0.15,0.7,0.15}
\definecolor{bag}{rgb}{0.6,0.6,0.2}

\newcommand{\lc}{LiDAR-camera~}

\newcommand{\lidar}{LiDAR~}

\newcommand{\lidarN}{LiDAR}
\newcommand{\lidars}{LiDARs~}
\newcommand{\lidarsN}{LiDARs}
\newcommand{\velodyne}{\textit{32-Beam Velodyne ULTRA Puck LiDAR~}}
\newcommand{\velodyneN}{\textit{32-Beam Velodyne ULTRA Puck LiDAR}}

\newcommand{\lidart}{LiDARTag~}
\newcommand{\lidarts}{LiDARTags~}
\newcommand{\lidartN}{LiDARTag}
\newcommand{\lidartsN}{LiDARTags}

\newcommand{\atag}{AprilTag~}

\newcommand{\atagN}{AprilTag}
\newcommand{\atagsN}{AprilTags}

\newcommand{\squeezeup}{\vspace{-4mm}} 

\newcommand{\overtilde}[1]{\mkern 1.5mu\widetilde{\mkern-1.5mu#1\mkern-1.5mu}\mkern 1.5mu}

\def\realnumbers{\mathbb{R}}
\def\real{\mathbb{R}}
\def\reals{\mathbb{R}}

\newcommand{\tp}{\mathcal{TP}}

\newcommand{\calx}{\mathcal{X}}

\DeclareDocumentCommand{\RI}{ O{} O{} }{\mathcal{RI}_{#1}^{#2}}

\newcommand{\floor}[1]{\lfloor #1 \rfloor}

\DeclareDocumentCommand{\td}{ O{} }{\tilde{#1}}
\newcommand{\transpose}{\mathsf{T}}

\newcommand{\SO}{\mathrm{SO}}

\newcommand{\SE}{\mathrm{SE}}

\newcommand{\Log}{\mathrm{Log}}

\DeclareDocumentCommand{\asin}{ O{} }{\sin^{-1}(#1)}
\DeclareDocumentCommand{\acos}{ O{} }{\cos^{-1}(#1)}
\DeclareDocumentCommand{\atan}{ O{} }{\tan^{-1}(#1)}

\DeclareDocumentCommand{\vector}{ O{} }{\mathrm{vec}(#1)}
\DeclareDocumentCommand{\zeros}{ O{} }{\textbf{0}_{#1}}
\DeclareDocumentCommand{\pre}{ O{} O{} }{{}_{#1}^{#2}}

\DeclareMathOperator*{\argmin}{arg\,min}

\newcommand{\Ical}{\mathcal{I}}

\newcommand{\Xcal}{\mathcal{X}}

\newcommand{\Zcal}{\mathcal{Z}}

\DeclareDocumentCommand{\A}{ O{} O{} }{\textbf{A}_{#1}^{#2}}
\DeclareDocumentCommand{\H}{ O{} O{} }{\textbf{H}_{#1}^{#2}}
\DeclareDocumentCommand{\I}{ O{} O{} }{\textbf{I}_{#1}^{#2}}
\DeclareDocumentCommand{\L}{ O{} O{} }{\textbf{L}_{#1}^{#2}}
\DeclareDocumentCommand{\M}{ O{} O{} }{\textbf{M}_{#1}^{#2}}
\DeclareDocumentCommand{\N}{ O{} O{} }{\textbf{N}_{#1}^{#2}}
\DeclareDocumentCommand{\P}{ O{} O{} }{\textbf{P}_{#1}^{#2}}
\DeclareDocumentCommand{\R}{ O{} O{} }{\textbf{R}_{#1}^{#2}}
\DeclareDocumentCommand{\T}{ O{} O{} }{\textbf{T}_{#1}^{#2}}
\DeclareDocumentCommand{\U}{ O{} O{} }{\textbf{U}_{#1}^{#2}}
\DeclareDocumentCommand{\V}{ O{} O{} }{\textbf{V}_{#1}^{#2}}
\DeclareDocumentCommand{\X}{ O{} O{} }{\textbf{X}_{#1}^{#2}}
\DeclareDocumentCommand{\Y}{ O{} O{} }{\textbf{Y}_{#1}^{#2}}

\DeclareDocumentCommand{\e}{ O{} O{} }{\textbf{e}_{#1}^{#2}}
\DeclareDocumentCommand{\n}{ O{} O{} }{\textbf{n}_{#1}^{#2}}
\DeclareDocumentCommand{\t}{ O{} O{} }{\textbf{t}_{#1}^{#2}}
\DeclareDocumentCommand{\p}{ O{} O{} }{\textbf{p}_{#1}^{#2}}
\DeclareDocumentCommand{\q}{ O{} O{} }{\textbf{q}_{#1}^{#2}}
\DeclareDocumentCommand{\u}{ O{} O{} }{\textbf{u}_{#1}^{#2}}
\DeclareDocumentCommand{\v}{ O{} O{} }{\textbf{v}_{#1}^{#2}}
\DeclareDocumentCommand{\x}{ O{} O{} }{\textbf{x}_{#1}^{#2}}

%% file: Introduction.tex
\section{Introduction}
\label{sec:intro}
%
%
%
Artificial landmark systems, referred to as \emph{fiducial markers}, have been designed for automatic detection via specific type of sensors such as cameras~\cite{wagner2003artoolkit,fiala2005artag,olson2011apriltag, wang2016apriltag,
krogiusflexible,degol2017chromatag}. The marker usually consists of a \emph{payload},
that is, a pattern that makes individual markers uniquely distinguishable, and a
boundary surrounding the payload that is designed to assist with isolating the
payload from its background. Their supporting algorithms typically consist of modules to detect the marker, decode its payload, and estimate the pose. Such artificial landmarks have been successfully used in
computer vision, augmented reality~\cite{wagner2003artoolkit} and simultaneous
localization and mapping (SLAM)~\cite{degol2018improved}. 

\begin{figure}[t]%
    \centering
    \subfloat{%
        \label{fig:first_image1}%
    \includegraphics[trim=100 100 0 150,clip,width=0.9\columnwidth]{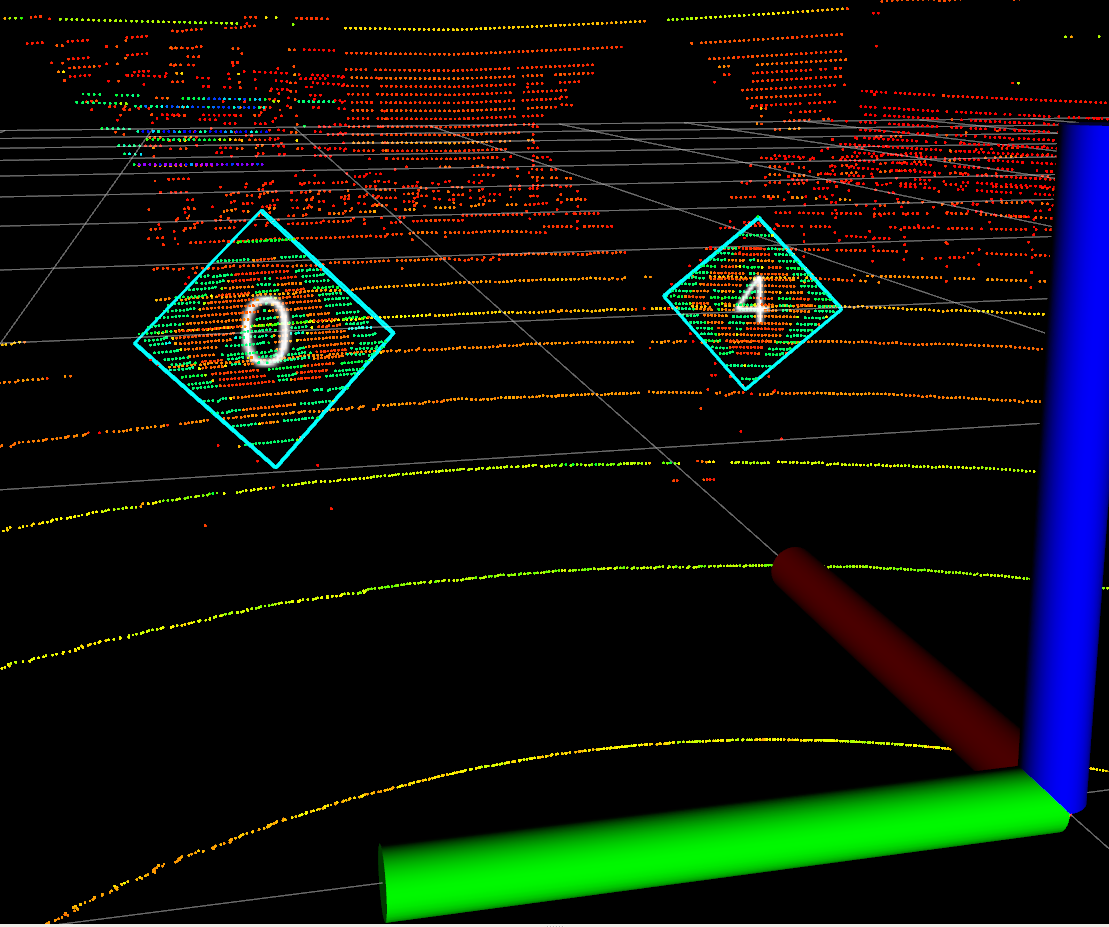}}\\
        \vspace{-3mm}
        \subfloat{%
        \label{fig:first_image2}%
    \includegraphics[trim=10 0 25 0,clip,width=0.9\columnwidth]{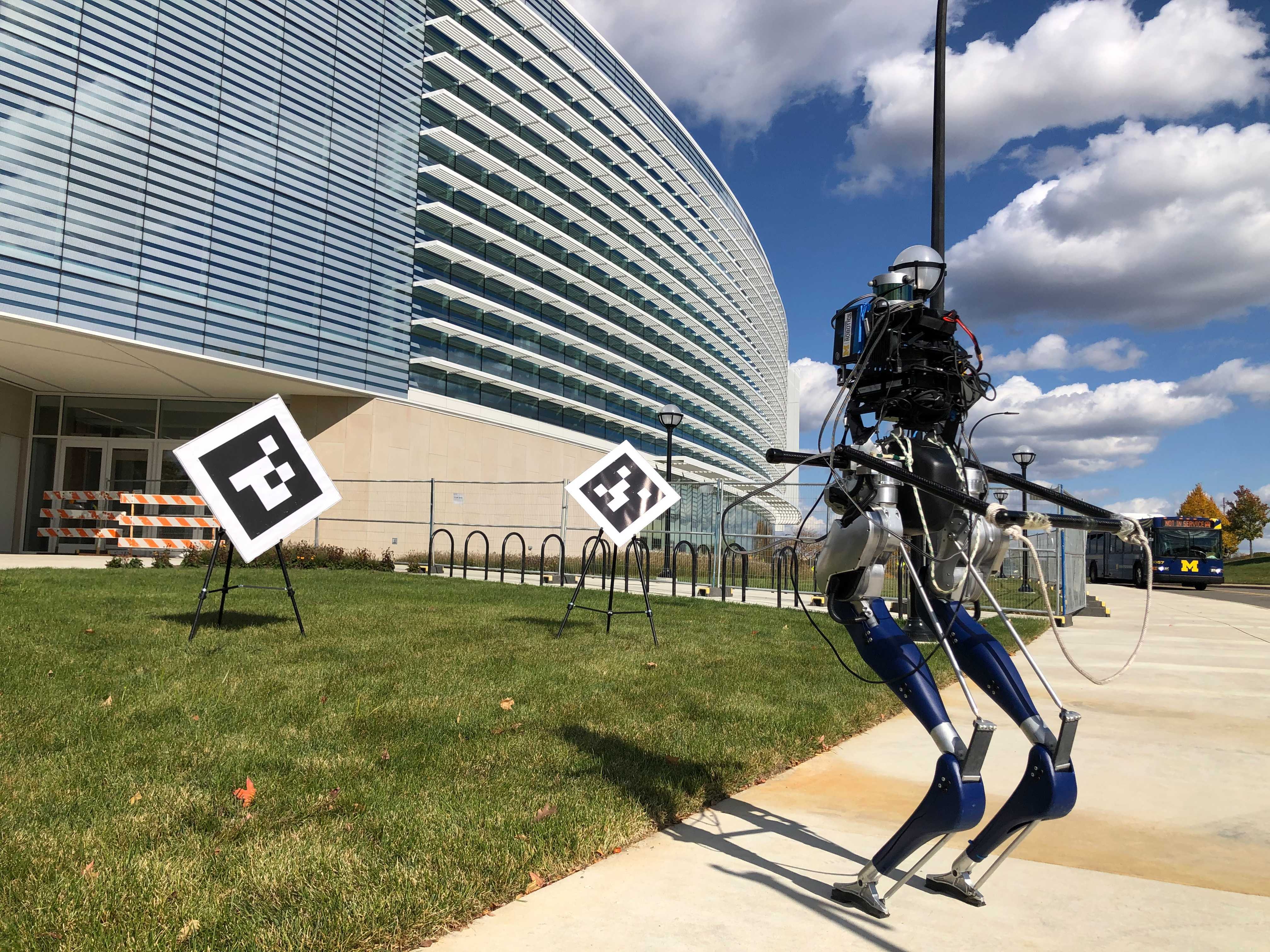}}%
    \caption[]{
        LiDAR-based markers can be used in tandem with camera-based markers to address the issue of images being sensitive to ambient lighting. 
        This figure shows a visualization of  \lidarts of two different sizes in a full point cloud scan.
}%
\squeezeup
\vspace{-3mm}
\label{fig:first_image}%
\end{figure}

Images are sensitive to lighting variations, and therefore, visual fiducial markers
rely heavily on the assumption of illumination consistency. As such, when lighting
changes rapidly throughout a scene, the detection of visual markers can fail.
Alternatively, light detection and ranging devices (\lidarsN) are robust to
illumination changes due to the active nature of the sensor. In particular, rapid
changes in the ambient lighting do not affect the detection of features in point
clouds returned by a \lidarN. Unfortunately, however \lidars cannot detect the fiducial
markers designed for cameras. Hence, to utilize the advantages of both sensor
modalities, a new type of fiducial marker that can be perceived by both \lidars and
cameras is required. Such a new marker can enable applications such as multi-sensor
fusion and calibration tasks involving visual and \lidar
data~\cite{huang2019improvements}. 
Designing such fiducial markers is challenging due to inherent \lidar
properties such as sparsity, lack of structure and the varying number of points in a
scan. In particular, the fact that an individual \lidar return has no fixed spatial
relation to neighboring returns makes it difficult to isolate a fiducial marker
within a point cloud.

In this paper, we propose a novel and flexible design of a fiducial tag system, called
\lidartN, as shown in Figure~\ref{fig:first_image}. A system is further developed to
detect \lidarts with various sizes and to estimate their poses. Point clouds are
represented as functions in a Reproducing Kernel Hilbert Space
(RKHS)~\cite{MGhaffari-RSS-19} to decode their IDs. The
whole system can run in real-time (over 100 Hz), which is even faster than currently
available data rates of \lidar sensors. The proposed \lidart can be perceived by both
RGB-cameras and point clouds. In~\cite{huang2019improvements,githubFileExtrinsic}, the authors used \lidarts in a \lc calibration pipeline as a means to extract the \lidar returns on the tag. The \lidart system was not used to decode a payload nor to estimate pose. \lidarts have been successfully used for \lc
calibration~\cite{huang2019improvements,githubFileExtrinsic}. 
It can be further applied to SLAM systems for robot state estimation and loop closures.
Additionally, it can help improve human-robot interaction, allowing humans to give
commands to a robot by showing an appropriate LiDARTag. However, the proposed system utilizes the intensity measurement of a \lidart to decode its ID. Therefore, \lidars with stable (good) intensity readings are required. In particular, the present
work has the following contributions:
\begin{enumerate}
    \item We propose a novel and flexible fiducial marker for point clouds, \lidartN, that is compatible with existing image-based fiducial marker systems, such as \atagN. 
    \item We develop a robust real-time method to estimate the pose of a \lidartN.
        The optimal pose estimate minimizes an $L_1$-inspired fitting error between
        the point cloud and the marker's template of known geometry. 
    \item To address the sparsity of \lidar returns, we lift a point cloud to a continuous function in an RKHS and use the inner product structure to determine a marker's ID among a pre-computed function dictionary.
    \item We present performance evaluations of the \lidart where ground truth data
        are provided by a motion capture system. We also extensively analyze each
        step in the system with spacious outdoor and cluttered indoor environments.
        Additionally, we report the rate of false positives validated on the indoor Google Cartographer~\cite{hess2016real} dataset and the outdoor Honda H3D dataset~\cite{360LiDARTracking_ICRA_2019}. 
    \item We provide open-source implementations for the physical design of the
        proposed \lidart and all of the associated software for using them, in C++
        and Robot Operating System (ROS)~\cite{quigley2009ros}; see
        \href{https://github.com/UMich-BipedLab/LiDARTag}{https://github.com/UMich-BipedLab/LiDARTag}~\cite{githubLiDARTag}.
\end{enumerate}

The remainder of this paper is organized as follows. Section~\ref{sec:related_work}
presents a summary of the related work. Section~\ref{sec:tag_design} explains the tag
design. Tag detection and pose estimation are discussed in
Section~\ref{sec:Detection} and Section~\ref{sec:PoseEstimationAndInitialization}.
The construction of continuous functions and ID decoding are introduced in
Section~\ref{sec:Decoding}. Experimental evaluations of the proposed \lidarts are
presented in Section~\ref{sec:results}. Finally, Section~\ref{sec:conclusion}
concludes the paper and provides suggestions for future work.

%% file: Related.tex
\section{Related Work}
\label{sec:related_work}
Fiducial marker systems were originally developed and used for augmented reality
applications~\cite{wagner2003artoolkit,fiala2005artag} and have been widely used for object detection and tracking and pose estimation~\cite{klopschitz2007automatic}. Due to their
uniqueness and fast detection rate, they are also often used to improve Simultaneous Localization And Mapping (SLAM) systems~\cite{degol2018improved}. Because of automatic detection, the camera-based markers are often used to extract features for cameras in target-based \lc calibration~\cite{huang2019improvements, Velas2014CalibrationOR, park2014calibration}, and the proposed algorithm can provide a means to extract features for \lidarsN. To the best of our knowledge, there are no existing fiducial markers for point clouds. Among the many popular camera-based fiducial markers are ARToolKit~\cite{wagner2003artoolkit}, ARTag~\cite{fiala2005artag}, AprilTag 1-3~\cite{olson2011apriltag,wang2016apriltag,krogiusflexible}, and CALTag~\cite{atcheson2010caltag}. 

In the following, we review some recent and well-known fiducial markers for cameras.
ARTag~\cite{fiala2005artag}~\cite{fiala2005comparing} uses a 2D barcode to make
decoding easier. \atag 1-3~\cite{olson2011apriltag, wang2016apriltag,
krogiusflexible} introduced a lexicode-based~\cite{trachtenbert1996computational} tag
generation method in order to reduce false positive detection.
ChromaTag~\cite{degol2017chromatag} proposes color gradients to speed up the
detection process. RuneTag~\cite{bergamasco2011rune} uses rings of dots to improve
occlusion robustness and more accurate camera pose estimation.
CCTag~\cite{calvet2016detection} adopts a set of rings to enhance blur robustness.
More recently, LFTag~\cite{wang2020lftag} has taken advantage of topological
markers, a kind of uncommon topological pattern, to improve longer detection
range. This also enables the decoding of markers with high distortion, and these markers can
be flexibly laid down. While there are some fiducial markers using deep learning
technique~\cite{grinchuk2016learnable, hu2019deep}, to date, all of those detectors,
still only work on cameras.

There are several deep-learning-based object detection architectures for \lidar point
cloud. Most of the methods for 3D object detection deploy a voxel grid representation
\cite{song2014sliding, engelcke2017vote3deep, zhou2018voxelnet}. Recently,
\cite{song2016deep, engelcke2017vote3deep, li20173d} have sought to improve feature
representation with 3D convolution networks, which require expensive computation.
Similar to proposed methods for 2D objects~\cite{zitnick2014edge,
van2011segmentation, carreira2011cpmc, li20203d, shi2019pointrcnn}, the proposed
methods for 3D objects generate a set of 3D boxes in order to cover most of the
objects in 3D space. However, most detectors are limited to specific categories and
none of these detectors or proposed methods has adequately addressed rotation,
perspective transformations, or domain adoption. 

\begin{remark}
As mentioned above, there exist several deep-learning-based object detectors trained
on large-scale \lidar datasets~\cite{Geiger2013IJRR, hackel2017isprs, kim2019pedx}.
These detectors are trained on limited categories in a specific dataset, and if the
training and testing data are not consistent, the inference process could fail. They
would have to be retrained on new data in order to be viable for our \lidartN.
Another option could be to design our own detector and train on our own datasets
rather than using existing detectors. However, there is no guarantee that the
resulting detector would work for varied scenes spanning from a cluttered laboratory to
spacious outdoor environments. Additionally, the existing detectors rely on powerful
graphics processing units (GPUs) and they are thus not suitable for lightweight
mobile robots. On the other hand, the detector proposed in this paper is robust to
general scenarios and achieves satisfactory results in practice. Deep-learning-based
methods, however, are interesting future work and are discussed in
Sec.~\ref{sec:conclusion}.
\end{remark}

%% file: TagDesign.tex
\section{Tag Design and \lidar Characteristics}
\label{sec:tag_design}
This section describes some essential points to consider when designing and using a
LiDAR-based tag system. In particular, this section addresses
how the unstructured point cloud from a LiDAR results in different considerations in
the selection of a marker versus those used with a camera. 

\subsection{LiDAR Point Clouds vs Camera Images}
\label{sec:lidarVScamera}
\begin{figure}[t]%
\centering
\subfloat{%
\includegraphics[height=0.45\columnwidth]{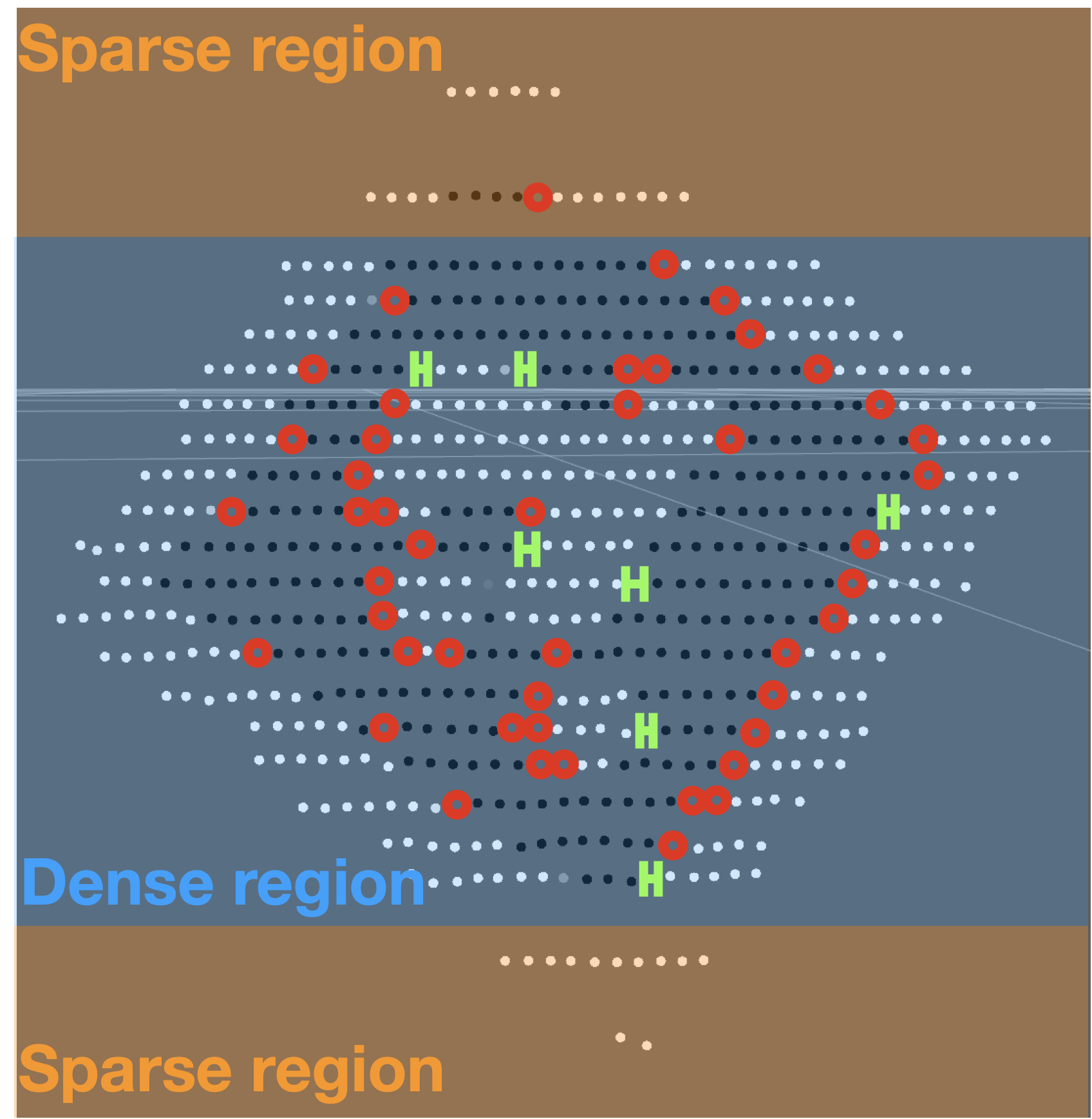}}%
\subfloat{%
\hspace{8pt}
\includegraphics[trim=0 0 0 0,clip, height=0.45\columnwidth]{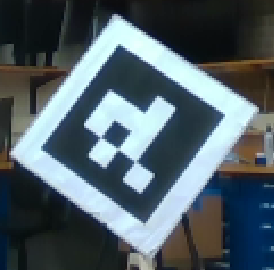}}%
\caption[]{This figure illustrates the unstructured nature of a LiDAR-point-cloud
return (left) for a planar surface with black and white squares (right). On the left,
the black and white dots are lower reflectivity and higher reflectivity, respectively.
The sparse region of the \lidar is indicated in light yellow and the dense region is
marked in light blue. The returns are more irregular at the black-white transitions.
Red circles indicate missing returns and green bars highlight larger gaps between
returned points.
}%
\label{fig:irregular}%
\end{figure}

Pixel arrays (i.e., an image) from standard RGB-cameras of different resolutions are
very structured, with the pixels arranged in a uniform (planar) grid, and each image
having a fixed number of data points. A LiDAR returns $(x,y,z)$ coordinates of 3D points lying on the surface of objects as well as the intensity, which
relies on the reflectivity/material of the object. The reflectivity is measured at the wavelength used by the \lidarN. Some \lidars also provide beam
numbers. The resulting 3D point clouds are typically referred to as ``unstructured''
because: 
\begin{itemize}
\item The number of returned points varies for each scan and for each beam. In
    particular, LiDAR returns are not uniformly distributed in angle or
    distance\footnote{Some \lidars have different ring density at different
    elevation angles. For example, \velodyne has dense ring density between
    $-5^\circ$ and $3^\circ$, and has sparse ring density from $-25^\circ$ to
    $-5^\circ$ and from $3^\circ$ to $15^\circ$~\cite{velodyneUltraPuck}.
    Therefore, in a sparse region, a target may be only partially 
    illuminated/observed.}.
\item  As shown in Fig.~\ref{fig:irregular}, high contrast between adjacent regions of a
    target's surface can result in missing returns and in varying spaces between
    returns. 
\item When used outdoors, the number of returned points is also
    influenced by environmental factors such as weather, especially temperature.
\end{itemize}
Consequently, as opposed to an image, there is no fixed geometric relationship
between the index numbers of returns from two different beams in a multi-beam LiDAR.
A further difference is the density of the collected data; currently, basic cameras
provide many more data points for a given surface size at a given distance than even
high-end \lidarsN. These summarized differences have an impact on how one approaches
the design of a LiDAR-based tag system vs. a camera-based tag system.

\subsection{Tag Placement and Design }
\label{sec:selection}
As mentioned in Sec. \ref{sec:lidarVScamera}, a return from a typical LiDAR
consists of an $(x,y,z)$ measurement, an intensity value $i$, and a
beam number (often called ring number, $r$). In this
paper, we propose exploiting the
relative accuracy of a LiDAR's distance measurement to determine ``features'' when
seeking to isolate a LiDARTag in a 3D point cloud (see Sec.~\ref{sec:Detection}) and
associating the isolated \lidart points with a
continuous function to decode the marker's payload (see Sec.~\ref{sec:Decoding}).

\subsubsection{Tag design}
\begin{figure}[t]%
\centering
\subfloat[]{%
    \label{fig:3dtag}%
    \includegraphics[height=0.45\columnwidth]{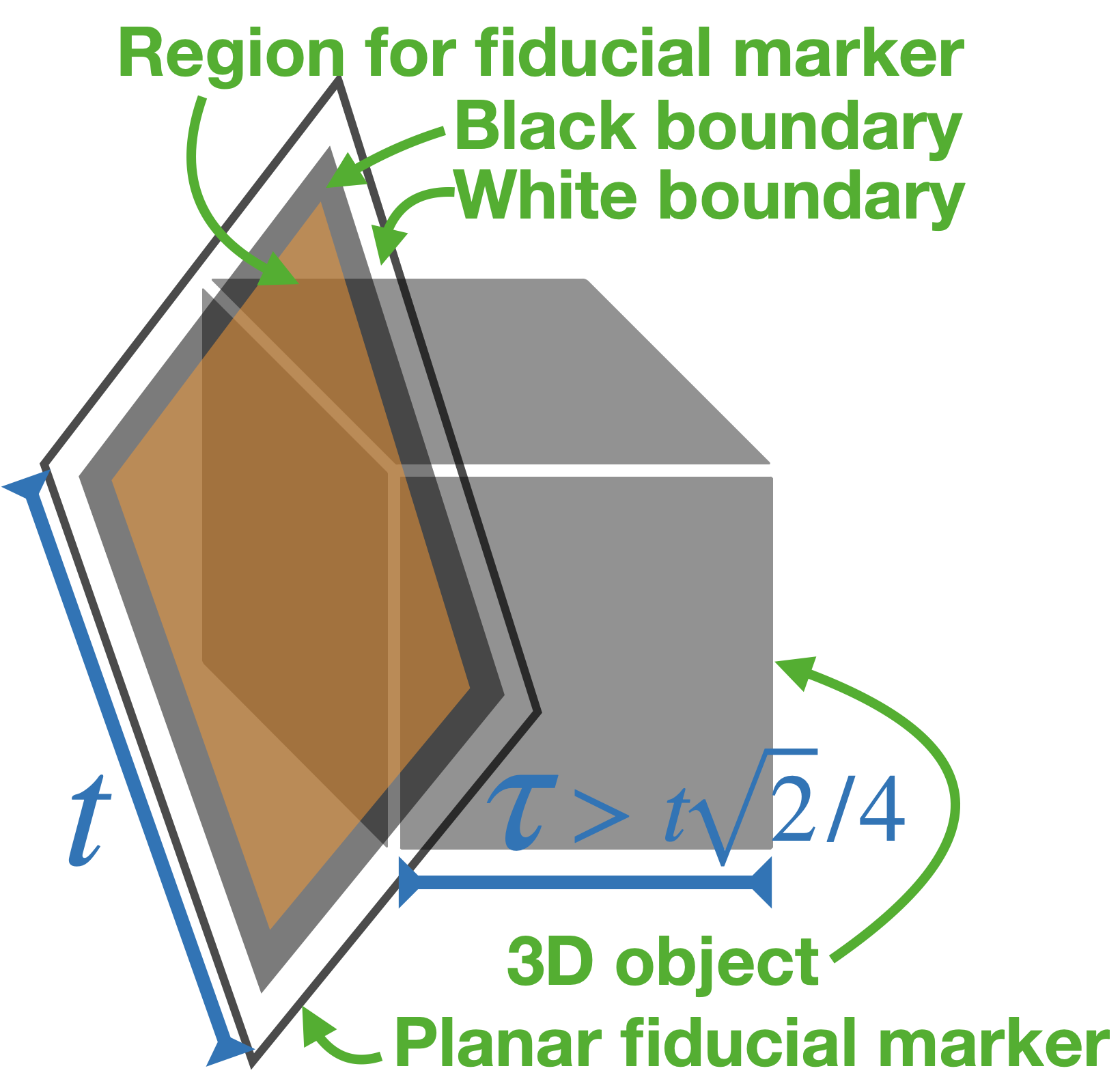}}%
\hspace{10pt}%
\subfloat[]{%
    \label{fig:tag_explain}%
    \includegraphics[height=0.45\columnwidth]{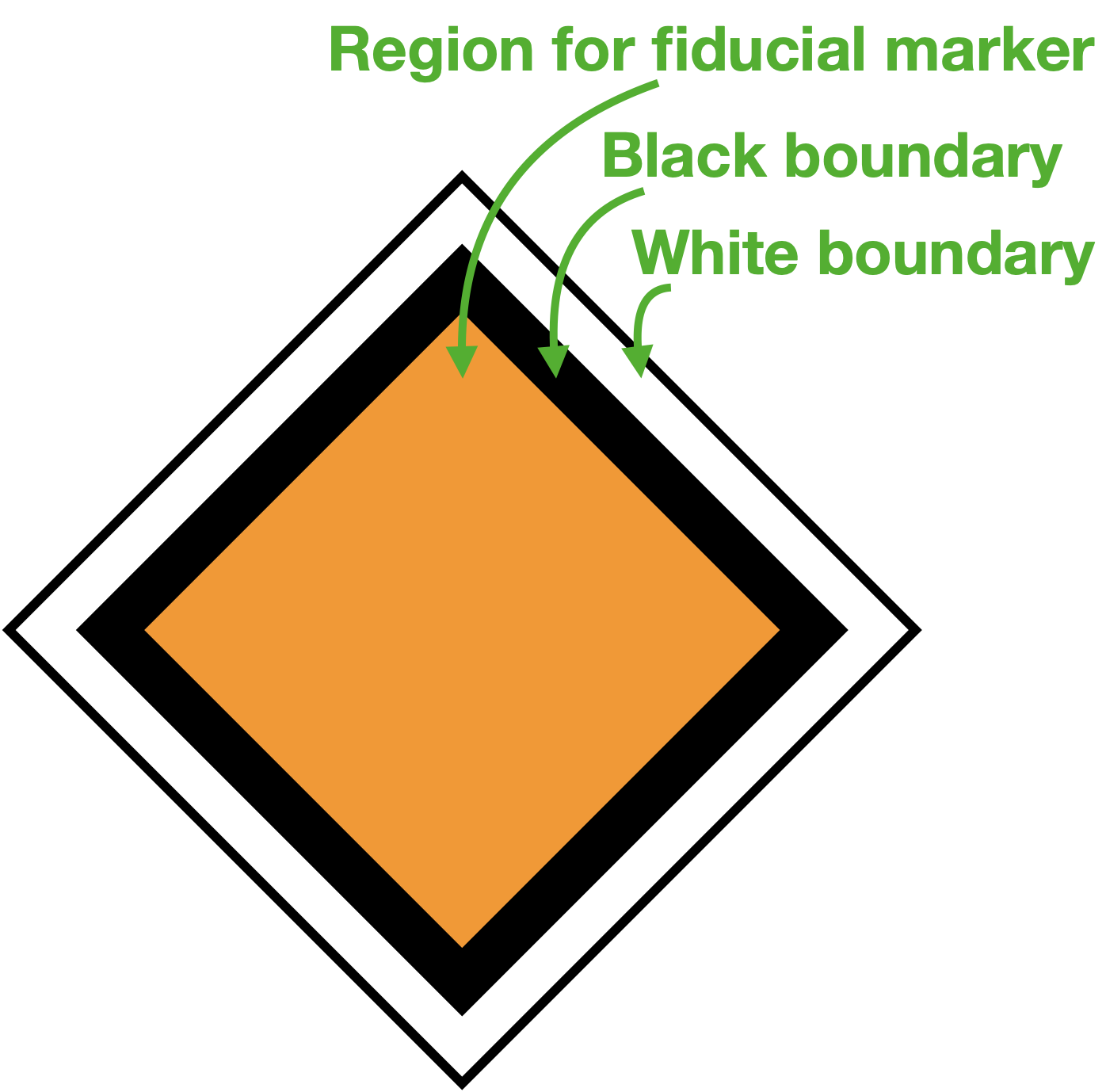}}\\
\subfloat[]{%
    \label{fig:MarkerExample}%
    \includegraphics[height=0.45\columnwidth]{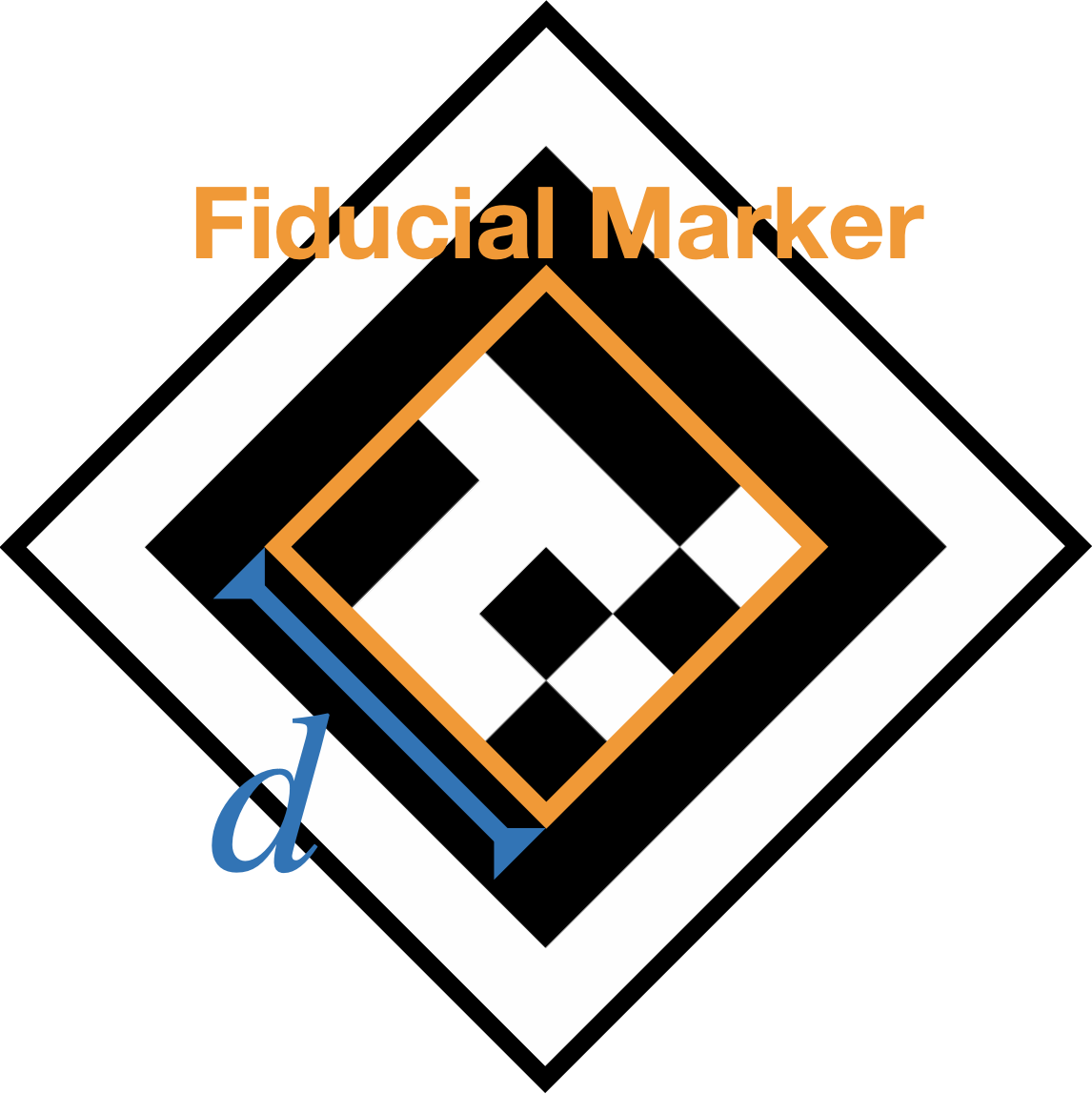}}%
\hspace{10pt}%
\subfloat[]{%
\label{fig:torso}%
\includegraphics[trim=0 10 0 0,clip,height=0.40\columnwidth]{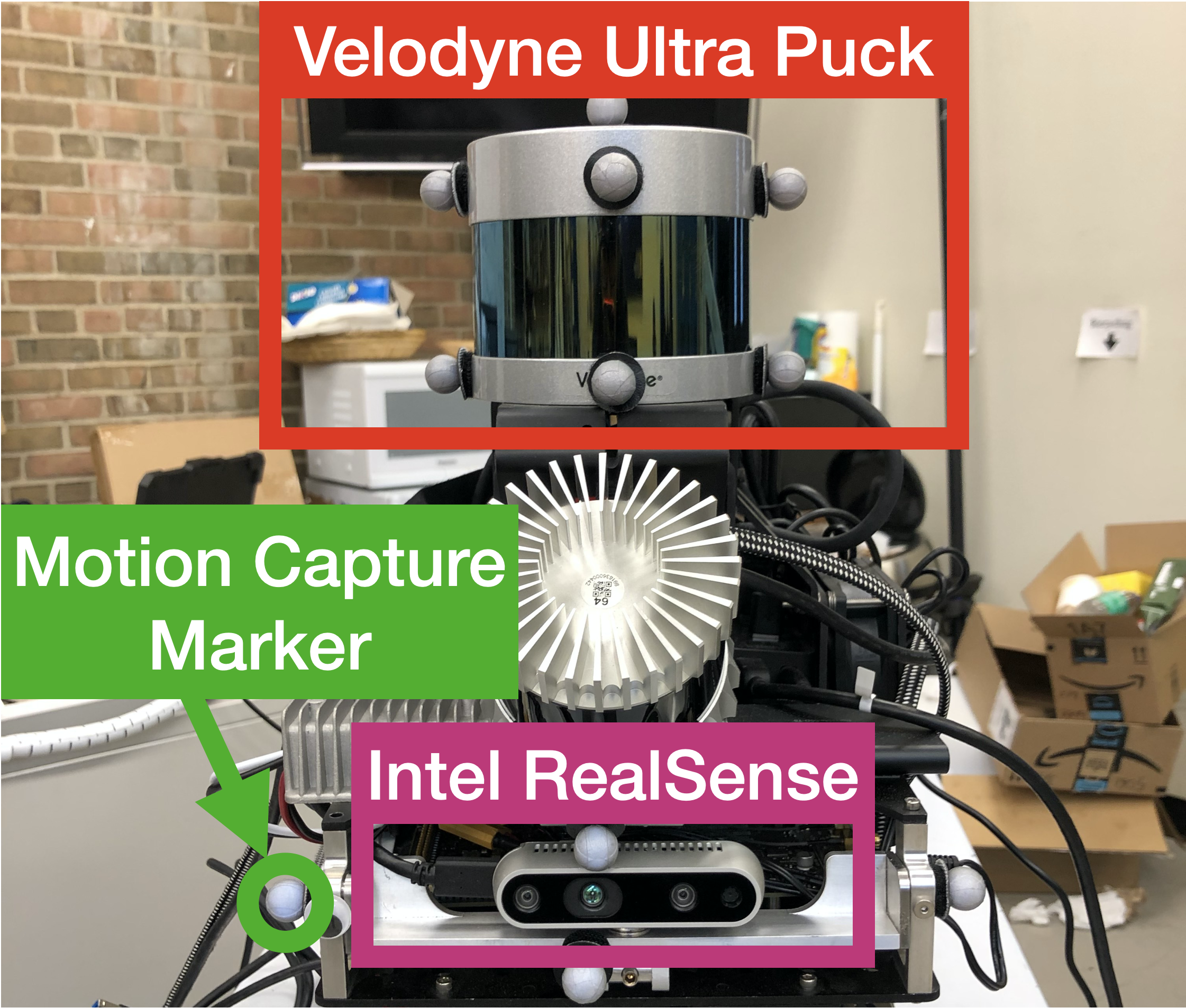}}%
\caption[]{\subref{fig:3dtag} illustrates a \lidart consisting of two parts: 
a 3D object with a rigidly attached, planar fiducial marker where $t$ and $h$ are the
marker size and height of the object respectively. \subref{fig:tag_explain} shows the
marker should be placed inside the yellow region and \subref{fig:MarkerExample}
illustrates an example of \atag being used as a \lidartN. \subref{fig:torso}
is the sensor setup consisting of a \lidarN, a camera and several motion capture
markers.
}%
\squeezeup
\label{fig:LiDARTag}%
\end{figure}

A LiDARTag is assumed to consist of a planar fiducial marker rigidly attached to
    a 3D object, as shown in Fig.~\ref{fig:3dtag}. In particular, the marker shows
different intensity values when illuminated by a \lidarN. As indicated in
    Sec.~\ref{sec:lidarVScamera}, intensity relies on how a \lidar measures the reflectivity of an object. Most types of fiducial markers for camera-based systems could be adapted for use in LiDAR-based systems as long as
    the payload is composed of differing reflectivities and is placed inside the
 region highlighted in yellow in Fig.~\ref{fig:tag_explain}. Figure~\ref{fig:MarkerExample} shows
    an example of an \atag used as a \lidartN. In our experiments, we print the tags from a regular printer and a poster machine. Because fiducial markers are usually
    printed from a printer, however, markers with two colors such as \atag 1-3
    \cite{olson2011apriltag, wang2016apriltag, krogiusflexible},
    ARTag~\cite{fiala2005artag, fiala2005comparing},
    InterSense~\cite{naimark2002circular}, CyberCode~\cite{rekimoto2000cybercode}, or
    CALTag~\cite{atcheson2010caltag} can be most easily adapted for use in our \lidart system, while
Cho et.al~\cite{cho1998multi} with multi-color cannot.

\begin{figure*}[t!]
    \includegraphics[width=\linewidth]{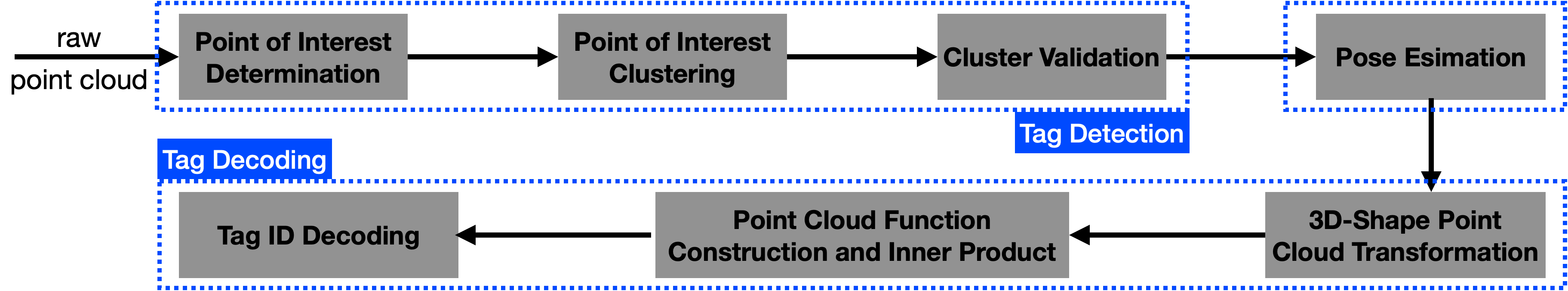}
    \caption{
        The system contains three parts: tag detection, pose estimation, and tag
        decoding. The detection step takes an entire \lidar scan (up to 120,000
        points from a \velodyneN) and outputs collections of likely payload points of
        the \lidartsN. Next, a tag's optimal pose minimizes the $L_1$-inspired cost
        in \eqref{eq:JKHcost}, though the rotation of the tag about a normal vector
        to the tag may be off by $\pm 90^\circ$ or $180^\circ$ and will be resolved in
        the decoding process. The tag's ID is decoded with a function library
        inspired by~\cite{clark2020nonparametric, ghaffari2019continuous}. The decoded
        tag removes the rotation ambiguity about the normal.}
    \label{fig:pipline}
    \squeezeup
\end{figure*}


For this initial study, we employ \atagN3 as our fiducial markers.
Furthermore, within the \atagN3 family of markers, we select tag16h6c5,
 that is, a tag encoding 16 bits (i.e., 16 black or white squares), with a
minimum Hamming distance of 6, and a complexity of 5. The Hamming distance measures
the minimum number of bit changes (e.g., bit errors) required to transform one string
of bits into the other. For example, the length-7 strings $``1011111"$ and
$``1001011"$ have a Hamming distance of 2, whereas $``1011111"$ and $``1001010"$ have
a Hamming distance of 3. The significance is that a lexicode with a minimum Hamming
distance $h$ can detect $h/2$ bit errors and correct up to $\floor{(h-1)/2}$ bit
errors~\cite{trachtenbert1996computational}. The complexity of an \atag is defined
as the number of rectangles required to generate the tag’s 2D pattern. For example, a
solid pattern requires just one rectangle, whereas, a white-black-white stripe would
need two rectangles (first draw a large white rectangle; then draw a smaller black
rectangle). For further details, we refer the reader to the coding discussion
in Olson~\cite{olson2011apriltag}.

\begin{remark}
One may argue that other members of the \atagsN3 family, such as tag49h15c15,
tag36h11c10, tag36h10c10, and tag25h10c8, would be more appropriate. For example,
including more bits tends to increase the number of distinct tags in a family, while
a larger Hamming distance reduces false positives. However, for tags of the same
physical size and the same distance from the sensor, more bits means fewer returns
per square and a higher error rate for individual squares.
\end{remark}

\subsubsection{Tag placement}
In Fig.~\ref{fig:3dtag}, let $t$ be the tag size and $h$ be the thickness of the 3D
object. The 3D object is assumed to have $t\sqrt{2}/4$ clearance around it, and
the first \lidar ring hitting at a \lidart is above $3/4$ of the \lidartN, see
Sec.~\ref{sec:clustering}. In particular, the fiducial marker can be
attached to a wall as long as the condition, $h>t\sqrt{2}/4$, in Fig.~\ref{fig:3dtag} is met. Finally, it is not recommended to orient the marker
like a square due to the quantization error inherited in \lidar sensors, see
\cite[Sec. II]{huang2019improvements}.


%% file: Detector.tex
\begin{figure}[t!]%
\centering
\subfloat[]{%
\label{fig:edges}%
    \includegraphics[height=0.23\textwidth]{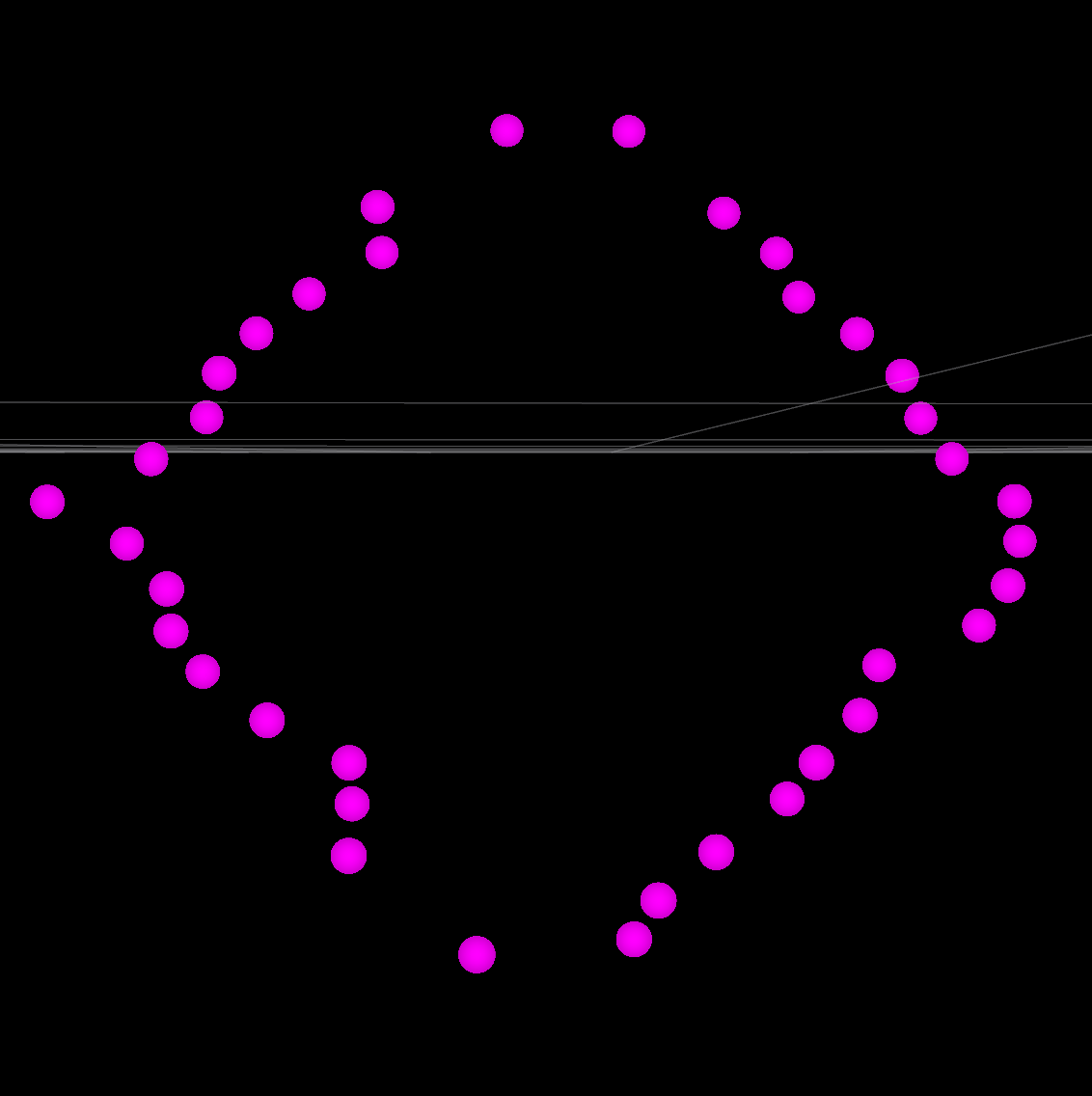}}~
\hspace{2pt}%
\subfloat[]{%
\label{fig:filled}%
    \includegraphics[height=0.23\textwidth]{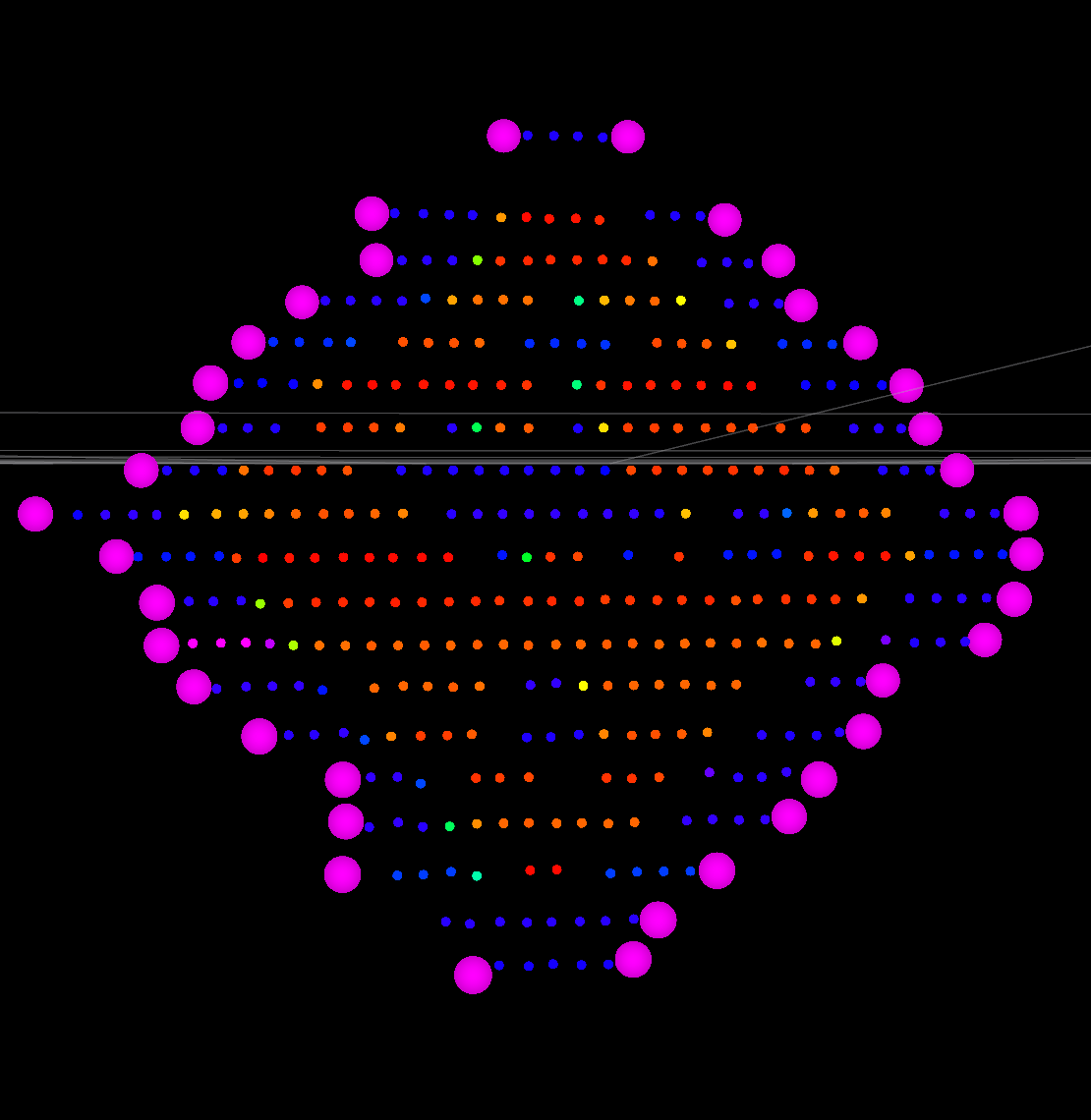}}\\
\hspace{2pt}%
\subfloat[]{%
\label{fig:boundary}%
    \includegraphics[height=0.23\textwidth]{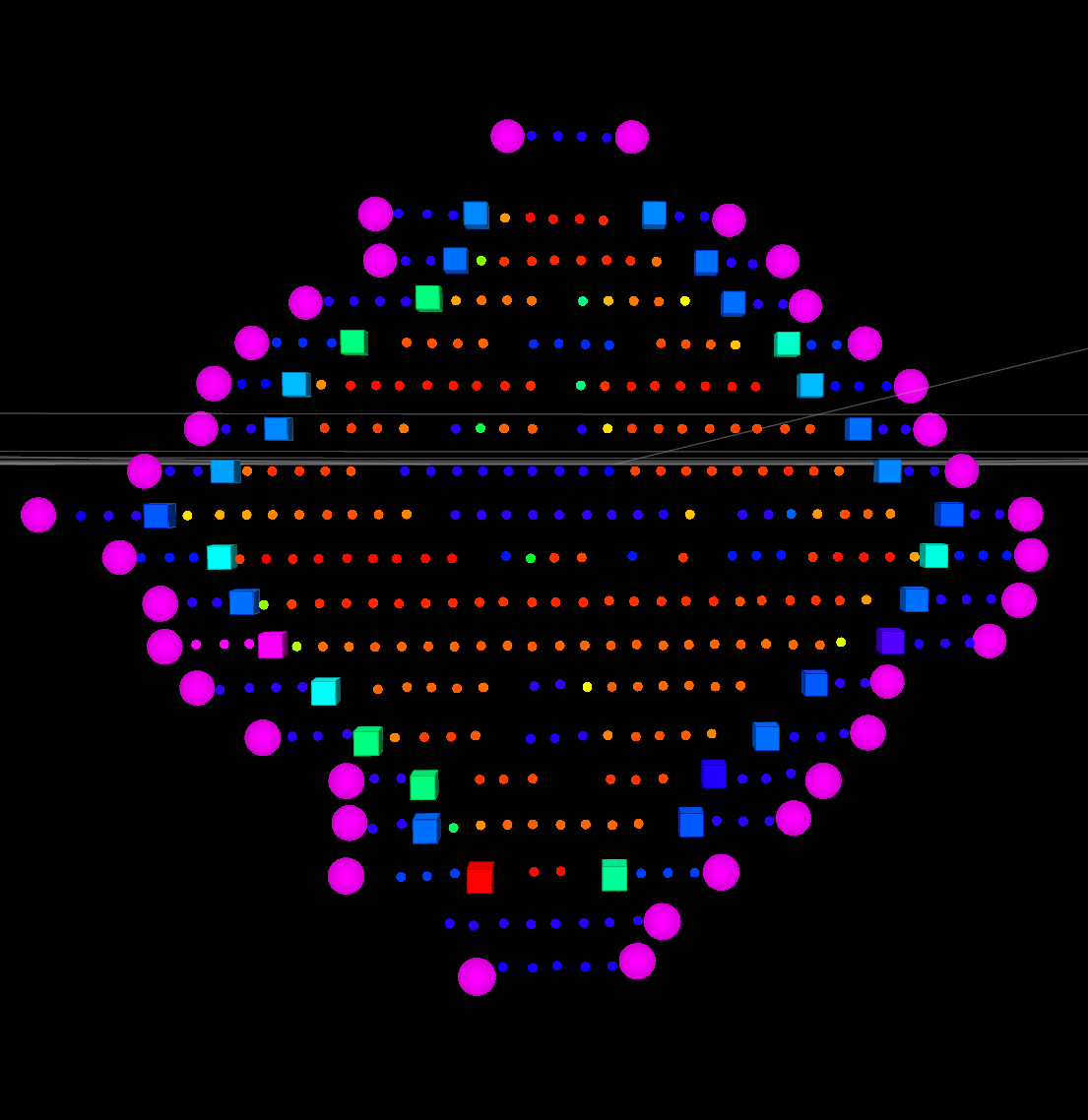}}~
\hspace{2pt}%
\subfloat[]{%
\label{fig:PlanarTemplate}%
    \includegraphics[trim=0 70 0 75,clip,height=0.23\textwidth]{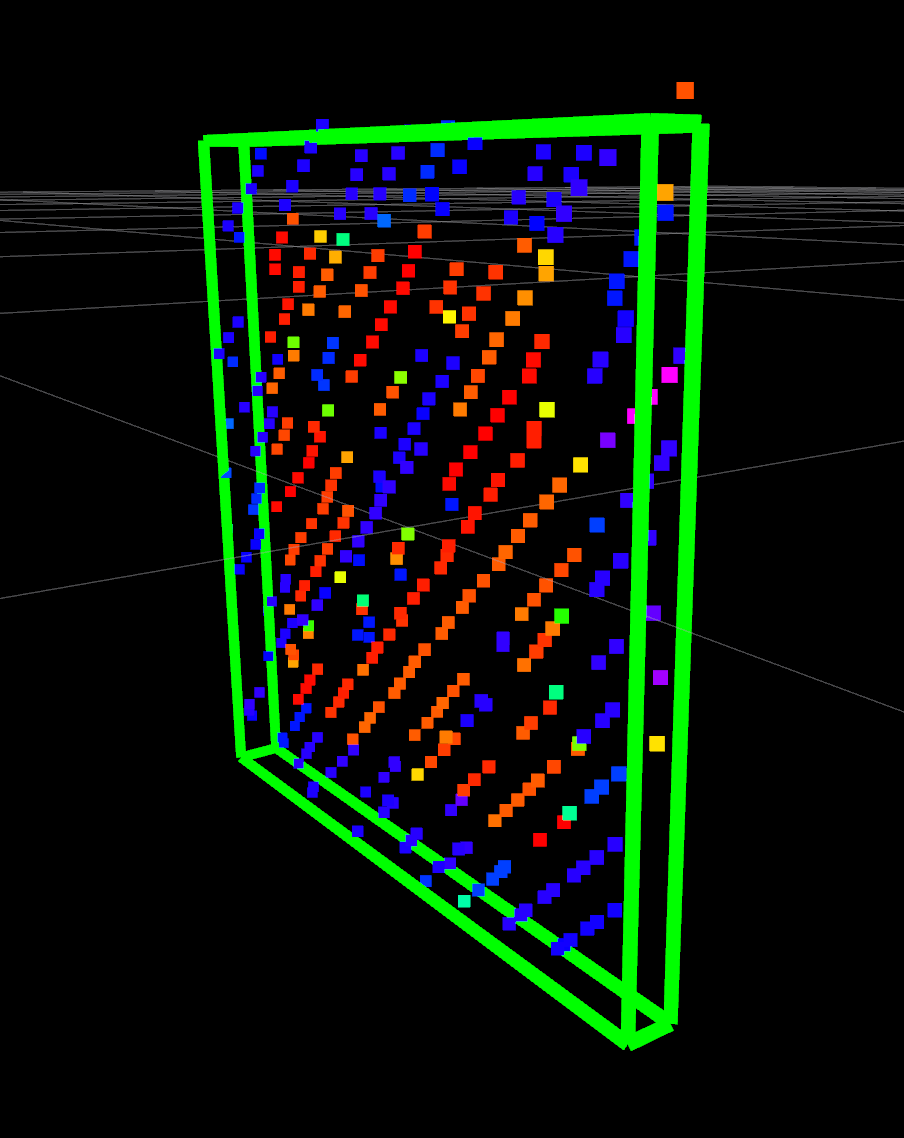}}~
\caption[]{
    Intermediate steps of the LiDARTag system. The system takes a full scan
point cloud and applies feature detection as defined in 
\eqref{eq:GD} and associates the features into clusters (magenta spheres), as shown
in~\subref{fig:edges}. Using the features, the clusters are filled from the
original scan (different color dots stand for different intensity values) as shown
in~\subref{fig:filled}. Later, boundary points (indicated by boxes) are detected.
After validating all the clusters, \subref{fig:PlanarTemplate} shows the result of
the point cloud of a \lidart
pulled back to the \lidar origin by $H_T^L$, which is a rigid-body
transformation from the tag to the \lidar that minimizes an $L_1$-inspired fitting
error \eqref{eq:JKHcost}. The green box is the template of the fiducial marker.
}%
 \label{fig:detector}%
\end{figure}

\section{Tag Detection}
\label{sec:Detection}
This section provides the individual steps for detecting potential
markers and examining their validity. The overall pipeline
is shown in Fig.~\ref{fig:pipline}. To localize potential \lidarts within the point
cloud, the first step is to find features. The features are then grouped
into distinct \emph{clusters}\footnote{Clustering features instead of clustering directly
on a \lidar scan is critical to achieve real-time applications.}. Most of these
clusters will not contain tags. Therefore, it is essential to validate whether a cluster contains a \lidart or not.

\subsection{Feature Detection}
\label{sec:edge}
As mentioned in Sec.~\ref{sec:lidarVScamera}, images are very structured in that the
vertical and horizontal pixel-pixel correspondences are known. Consequently, various
kinds of 2D kernels~\cite{canny1987computational} can be applied for edge detection.
However, unlike images, raw \lidar point clouds are unstructured in that even if we
have the indices of all points in each beam, we do not know the vertical point-point
correspondences. 

Therefore, to find a feature in a point cloud, an edge point is
defined as discontinuities in distance. Inspired by the point
selection method in LeGO-LOAM~\cite{legoloam2018}, a point is defined as a
feature if it is an edge point, and its consecutive $n$ points are not edge
points (similar to plane features in LeGO-LOAM). Given consecutive $n+1$ points, we
choose to use a 1D kernel to compute spatial gradients at each point to find edge
points. Let $p_{i,m}$ be the $i^{\text{th}}$ point in the $m^{\text{th}}$ beam so that the gradient of distance $\nabla D(p_{i,m})$ can be defined as
\begin{equation}
\label{eq:GD}
         \nabla D(p_{i,m}) = \|p_{i+l,m} - p_{i,m}\|_2 - \|p_{i-l,m} - p_{i,m}\|_2, 
\end{equation}
where $\ell$ is a design choice (here, $\ell=1$). If $\nabla D(p_{i,m})$ at a point
exceeds a threshold $\zeta$, we then consider $p_{i,m}$ as a possible edge point.
Using distance gradients requires a \lidart being $\zeta$
away from the background. For speed in real time application, we do not apply
further noise smoothing, edge enhancement, nor edge localization. Finally, if
there is only one edge point in the consecutive $n+1$ points ($n=2$ in this
paper), then the edge point is considered as a feature.

\subsection{Feature Clustering}
\label{sec:clustering}
\begin{figure}[t]%
\centering
\includegraphics[width=0.7\columnwidth]{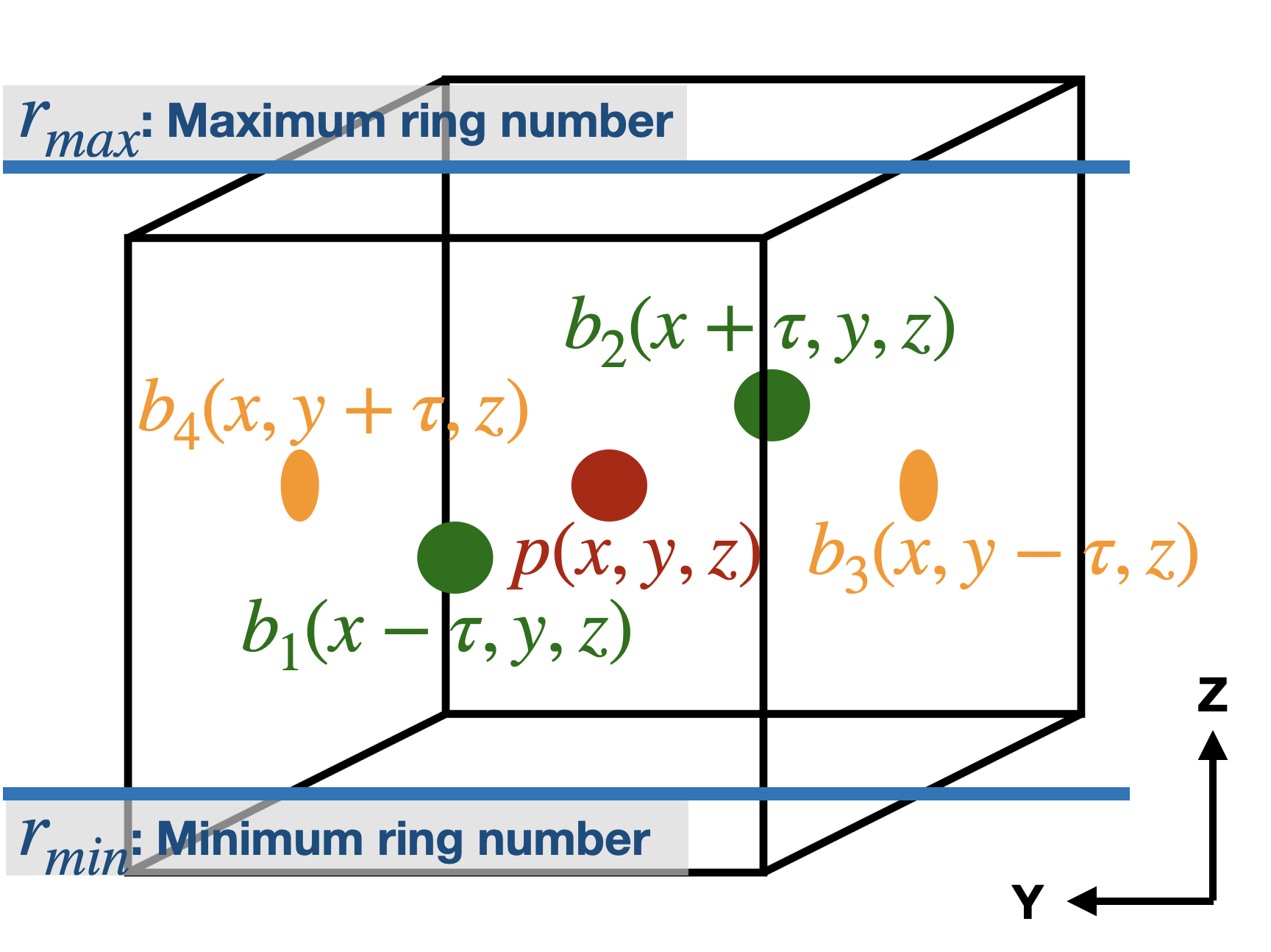}%
\caption[]{This figure shows the initial state of a cluster, in which has only single
    one feature. A cluster is defined as a cuboid in $\realnumbers^3$. When a feature
    fails at linkage, a cluster will be created and centered at itself
$(x,y,z)$ with four boundaries $(x\pm\tau, y\pm\tau, z)$ for the $x$-$y$ axes, where $\tau$ is
$t\sqrt{2}/4$, as well as the maximum ring number and the minimum ring number for the
$z$-axis.
}%
\label{fig:cluster}%
\end{figure}

After determining the features in the current point cloud, we group them
into clusters using the single-linkage agglomerative hierarchical clustering
algorithm\footnote{We chose this clustering algorithm because the number of
        \lidarts is unknown. Therefore, algorithms like K-Means Clustering cannot be
used.}~\cite{johnson1967hierarchical, lloyd1982least}. As indicated in
Sec.~\ref{sec:lidarVScamera}, \lidar returns are not uniformly distributed in angles
or distance. The linkage criteria, therefore, considers the $x$-$y$ axes and the
$z$-axis differently: signed Manhattan distance is chosen for the $x$-$y$ axes, and
ring numbers are selected for the $z$-axis. Similarly, boundaries of a cluster are
defined by the four center points of a cuboid's faces and maximum/minimum ring
numbers, as shown in Fig.~\ref{fig:cluster}. The algorithm loops over each
feature, either linking it to an existing cluster and updating its boundaries,
or creating a new cluster. 

\begin{remark}
    \lidar rings are determined by the elevation angle of an emitter.
    Most existing \lidars provide not only $(x,y,z,i)$ values, but also a ring
    number of a data point. If ring numbers are not available and there exists only one
    rotation axis in the \lidarN, a ring number can be simply regressed against
    elevation angle by taking the \lidarN's ring numbers as a discrete set of
    corresponding elevation angles, in which the number of elevation angles is the
    same as the number of beams of the \lidarN. The elevation angle of a data point
    can be computed as 
    $$\arctan\left({\frac{z}{\sqrt{x^2+y^2}}}\right).$$
    We use
    this method to regress ring numbers for the Google Cartographer dataset. For more
    detail, see our implementation on GitHub~\cite{githubLiDARTag}.
\end{remark}

\begin{figure}[t]%
\centering
\subfloat[]{%
\label{fig:PoseIllustration}%
\includegraphics[width=0.45\columnwidth]{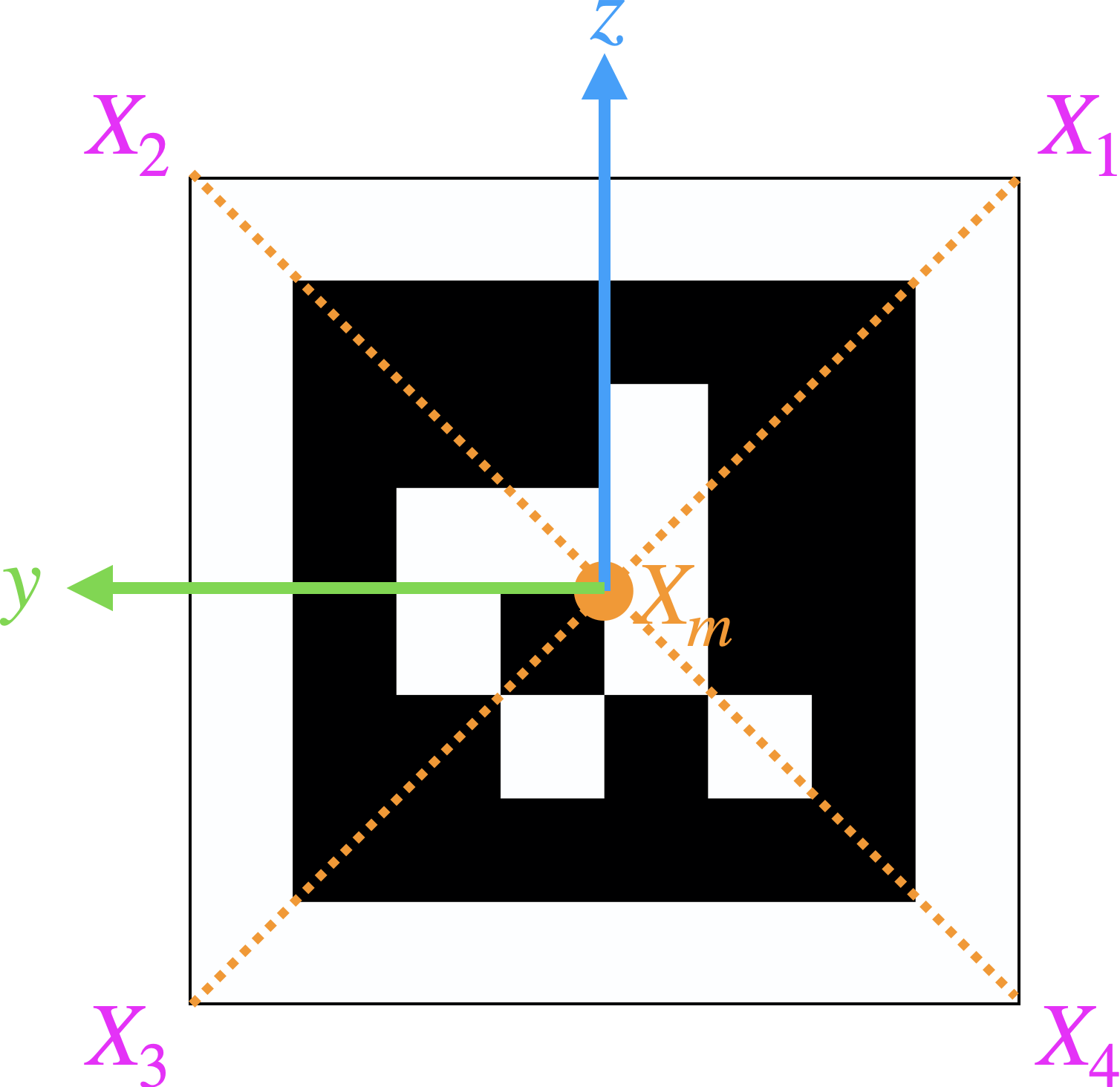}}%
\hspace{15pt}%
\subfloat[]{%
    \label{fig:TopMostBeam}%
\includegraphics[width=0.45\columnwidth]{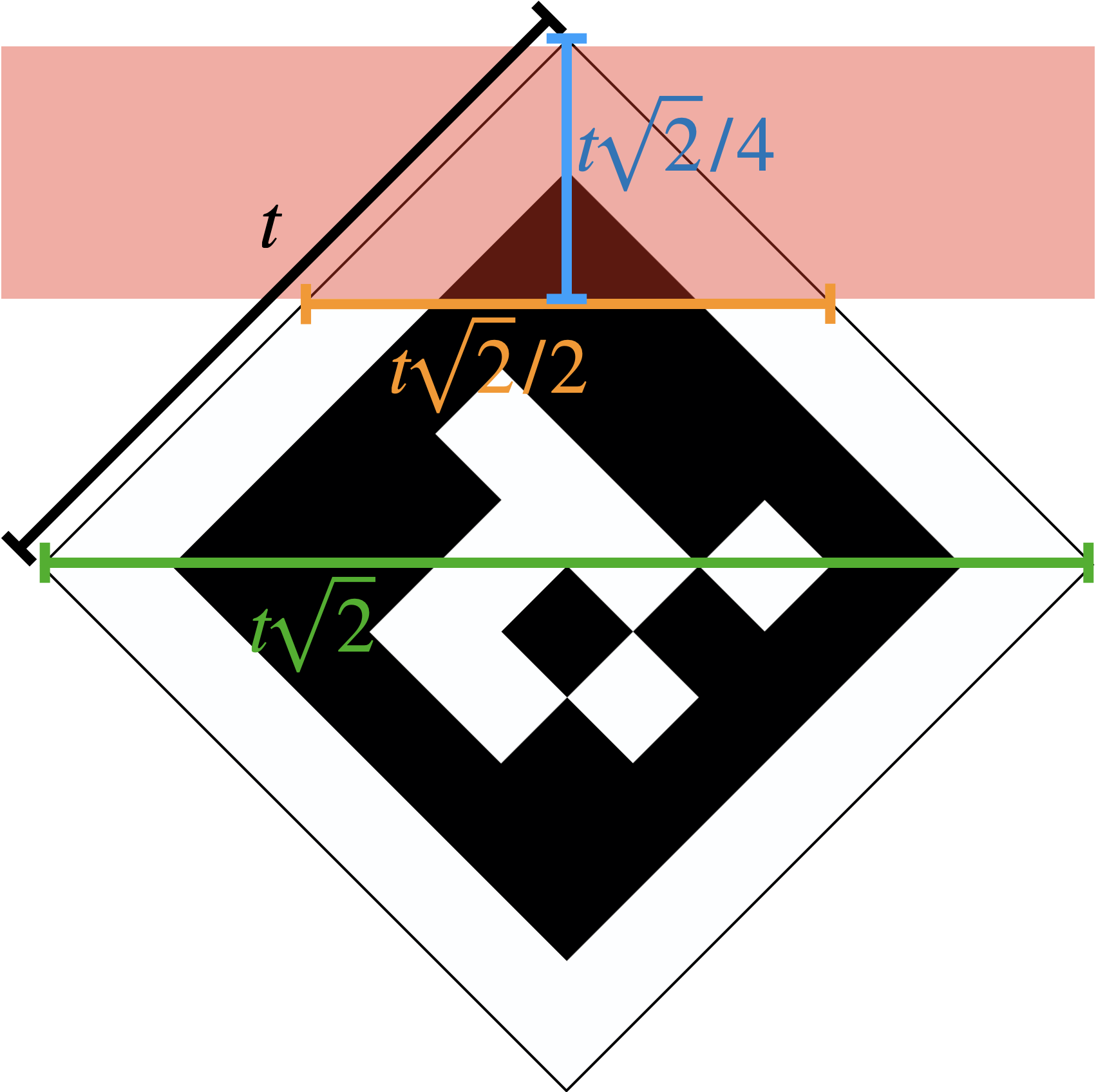}}%
\caption[]{\subref{fig:PoseIllustration} describes the coordinate system of the
    fiducial marker. \subref{fig:TopMostBeam} indicates that the first beam hitting the
\lidart should be at least $3/4$ above target, outlined as the red region.
}%
\end{figure}

The boundaries $(b_1,\cdots,b_4, r_{max}, r_{min})$ of a cluster are the
maximum/minimum $x$, maximum/minimum $y$ and maximum/minimum ring numbers among all
features in the cluster. When a new cluster is created from a feature
$p^k=(x, y, z, i, r)$, the four center points of the faces are defined as $(x\pm
\tau, y\pm \tau)$, with $\tau = t\sqrt{2}/4$ for the
$(b_1,\cdots,b_4)$. The $r_{max}$ and $r_{min}$ are defined in terms of the ring numbers $r$, as shown in
Fig.~\ref{fig:cluster}. The linkage criteria $L(p^k,c_j)$ between a feature
$p^k$ and a cluster $c_j$ is 
\begin{equation}
\begin{cases}
 \label{eq:linkage}
&\min (b_i^k - \tau)  \leq p^k \leq \max (b^k_i + \tau), ~\forall i=1, \cdots,4\\
&r_{min}-1  \leq r \leq r_{max}+1,
\end{cases}
\end{equation}
where the first line is the signed Manhattan distance for the $x$-$y$ axes and the second is the ring number for the $z$-axis. If both
conditions are met, the feature is linked to the cluster. The corresponding boundaries
are updated, if necessary. Due to this linkage criteria, a \lidart
requires $\tau = t\sqrt{2}/4$ clearance around it to
avoid false linkage, and based on preliminary testing, we also impose that the topmost beam on the \lidart should be above $3/4$ of
the target, as indicated by the red region in Fig.~\ref{fig:TopMostBeam}. 

\begin{remark}
    We chose not to use a k-d tree structure~\cite{bentley1975multidimensional}
    because the number of features and the resulting clusters are not large enough to
    benefit from the data structure. The construction time of the tree could overtake 
    the querying time.
\end{remark}

\subsection{Cluster Validation}
\label{sec:validation}
At this point, we have grouped the features into clusters as shown in
Fig~\ref{fig:edges}. In practice, few (possibly none) of the clusters will contain a
valid tag and thus it is important to be able to eliminate clusters that are clearly
invalid. To do so, we first used the point cloud data to fill in LiDAR returns between the features of a cluster, as
shown in Fig. \ref{fig:filled}. 

Inspired by AprilTag~\cite{olson2011apriltag,wang2016apriltag},
tag-family-based heuristics are used to validate a cluster: number of points
$\eta$ and number of features $\psi$ in the cluster. Another geometry-based heuristic is also
deployed: the outlier percentage $(\kappa)$ of a plane fitting process. To save
computation time, if any of the above processes fails, the cluster is marked as
invalid and does not proceed to the next stage of validation. The first two values are
determined by what type of tag family is chosen. Shown in Fig.~\ref{fig:MarkerExample}
is a tag family which contains 16 bits.

A lower bound on the number of points in a cluster is determined by how many bits
are contained in the fiducial marker. If the payload is $d \times d$, the LiDARTag
including boundaries is $(d+4) \times (d+4)$. If we assume a minimum of five returns
for each bit in the tag, then the minimum number points in a valid cluster is
\begin{equation}
    \eta \ge 5(d+4)^2.
\end{equation}

On the other hand, an upper bound is determined by the distance from the \lidar
to the marker and its size $t$. Given a tag at distance $D$, the maximum number of
returns on the marker happens when it directly faces to the \lidar and can be
computed as:
\begin{equation}
    M \frac{t\sqrt{2}}{D\sin{\theta}},
\end{equation}
where $\theta$ is the horizontal resolution of the \lidarN, $M$ is the number of
rings hitting on the tag, and $t\sqrt{2}$ is the diagonal length of the tag, as shown
in Fig.~\ref{fig:TopMostBeam}.

The boundaries of the payload can be detected by an intensity gradient, 
\begin{equation}
        \nabla I(p_{i,m}) = 
        |p_{i+\ell,m}^I - p_{i,m}^I| - |p_{i-\ell,m}^I - p_{i,m}^I|,
        \label{eq:GI}
\end{equation}
where $\ell$ is also a design choice (here, taken as one), and $p_{i,m}^I$ is the
intensity value of the $i^{\text{th}}$ point on the $m^{\text{th}}$ ring. An
example of detected boundary points is shown in Fig.~\ref{fig:boundary}.
If $\nabla
I(p_{i,m})$ exceeds a threshold, then $p_{i,m}$ is a payload edge point. To
successfully decode a tag, we will need at least one ring on each row of the payload.
Hence, the minimum number of payload edge points is
\begin{equation}
    \psi \ge 2(d+2).
    \label{eq:psi}
\end{equation}
Finally, we apply a plane fitting process to the remaining clusters. If the percentage
of outliers of the plane fitting is more than $\kappa$ (chosen as $0.05$), the
cluster is considered invalid.

The above heuristics allow us to extract a potential fiducial marker from the \lidart
through features in both cluttered indoor and spacious outdoor
environments. The next step is to estimate the pose of the marker.

\begin{figure}[t]%
\centering
\includegraphics[width=0.7\columnwidth]{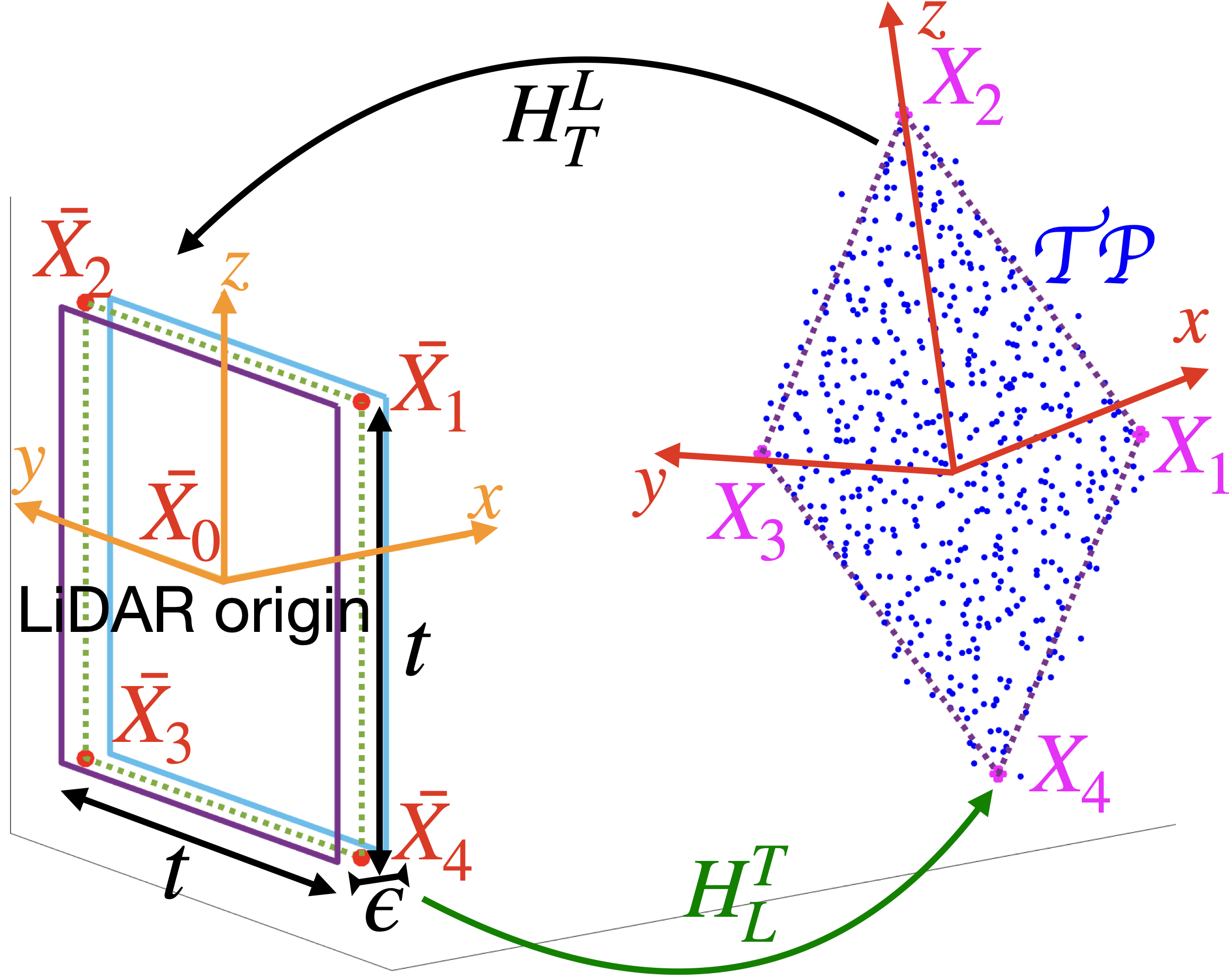}%
\caption[]{
This conceptual figure illustrates the proposed method to estimate a \lidartN's
pose. The target’s coordinate frame is defined as the mean of the four vertices 
$({X}_1, \cdots, {X}_4)$ and the template of known geometry is defined 
by $(\bar{X}_1, \cdots, \bar{X}_4)$ with depth $\epsilon$ at the \lidar origin.
The rigid-body transformation $H_T^L$ (black arrow) projects the target’s
point cloud to the template. The actual pose of the \lidart is estimated by
\eqref{eq:Pose} using the inverse transformation $H_T^L$ (green arrow).
}%
\label{fig:IdealFrame}
\end{figure}

%% file: PoseEstimation.tex
\section{Pose Estimation and Initialization}
\label{sec:PoseEstimationAndInitialization}
The pose of a \lidart is defined as $H_L^T$, a rigid-body transformation from the
\lidar frame to the \lidart frame, as shown in Fig.~\ref{fig:IdealFrame}. To estimate
the pose, we employ the $L_1$-inspired method proposed in
\cite{huang2019improvements}. The pose estimation is formulated into an optimization
problem~\eqref{eq:Pose} in Sec.~\ref{sec:PoseEstimation}. Due to $\SE(3)$ being
non-convex and the requirement for a fast estimate, initial guesses to initialize the
optimization problem and the gradient of the cost function are necessary, see
Sec.~\ref{sec:FormulationAndInitialization}.

\subsection{\lidart Pose Estimation}
\label{sec:PoseEstimation}
Define the target point cloud $\tp:= \{\Xcal_i\}_{i=1}^M$ as the collection of \lidar returns from a \lidartN,
where $M$ is the number of points. Given the target geometry, we define a template
with vertices $\{ \bar{X}_i \}_{i=1}^4$ located at the origin of the \lidar and aligned with the $y$-$z$ plane as
defined in Fig.~\ref{fig:IdealFrame}. 
We therefore seek a rigid-body transformation from \lidar to the tag, $H_L^T \in \mathrm{SE}(3)$, that ``best fits'' the template onto the \lidar returns of the target. In practice, it is actually easier to project the target point cloud
$\tp$ back to the origin of the LiDAR through the inverse of the current estimate of
transformation $H_T^L:=(H_L^T)^{-1}$ and measure the error there. The action of $H\in \mathrm{SE}(3)$ on $\reals^3$ is $H \cdot \Xcal_i = R \calx_i + p$, 
where $R\in\SO(3)$ and $p\in\reals^3$.
For $a \ge 0$ and
$\lambda \in \real$, an $L_1$-inspired cost is defined as
\begin{equation}
    \label{eq:L1cost}
    c(\lambda,a):=\begin{cases}
    \min\{ |\lambda-a|, |\lambda + a| \} & \text{if}~|\lambda| >a \\
    0 & \text{otherwise}
    \end{cases}.
\end{equation}
Let $\{\bar{\calx}_i\}_{i=1}^N:=H_T^L(\tp):=\{ H_T^L \cdot \calx_i \}_{i=1}^N$ 
denote
the projected point cloud by $H_T^L$, and denote a point's $(x,y,z)$-entries by
$(\bar{x}_i,\bar{y}_i,\bar{z}_i)$.
The total fitting error of the point cloud is defined as
\begin{equation}
	\label{eq:JKHcost}
    C(H_T^L(\tp)) :=\sum_{i=1}^{M} c(\bar{x}_i,\epsilon) +
    c(\bar{y}_i,d/2) +  c(\bar{z}_i,d/2),
\end{equation}
where $\epsilon \ge 0$ is a parameter to account for uncertainty in the depth
measurement of the planar target and the principal axis with the smallest
variance is used, see Sec.~\ref{sec:FormulationAndInitialization}. The optimization
problem becomes
\begin{equation}
    \label{eq:Pose}
    H_T^{L^*} := \argmin_{R_T^L, p_T^L} C(H_T^L(\tp)).
\end{equation}
Finally, the pose of a \lidart is $H_L^T = H_T^{L^*};$
see~\cite{huang2019improvements} for more details. To solve this optimization
problem, we leverage a gradient-based solver in the NLopt library~\cite{NLopt} and
the closed form of the gradient, which is provided on our
GitHub~\cite{githubLiDARTag}. 

Figure~\ref{fig:PlanarTemplate} shows the projected returns of a \lidart being
inside the green box (aligned with the $y$-$z$ plane) at the \lidar origin. In
addition, we further compute the 2D convex hull within the $y$-$z$ plane of the
pullback of point cloud and utilize the surveyor's formula~\cite{braden1986surveyor}
to calculate the area of the convex hull. Our assumption on where first ring hits the
marker results in at least 75\% of the marker's area being illuminated. Therefore, if
the estimated area is less than 75\% of the marker size, the cluster is considered
invalid.

\begin{remark}
    Equation \eqref{eq:Pose} provides an estimated rigid-body transformation from the \lidar to the tag,
    and importantly, due to the symmetric of the target, the rotation of the tag about a normal vector to the tag may be off by
    $\pm90^\circ$ or $180^\circ$.
    In particular, the four rotations in Fig.~\ref{fig:codeword_rotation} are not
    determined. This ambiguity will be removed after decoding the tag, see
Sec.~\ref{sec:Decoding}. 
\end{remark}

\subsection{Optimization Initialization} 
\label{sec:FormulationAndInitialization}
The corners of the \lidartsN, $({X_1}, \cdots,
{X_4})$, are estimated from $\tp$. The initial guess of a rigid-body transformation is chosen to minimize the distance
from the points $(\bar{X_1},\cdots, \bar{X_4})$ and $({X_1}, \cdots,
{X_4})$. This will be reduced to a (constrained orthogonal) procrustes
problem~\cite{gower1975generalized}, namely a problem of the form: 
\begin{equation}
\Theta = \argmin_{\substack{\Omega:\ \Omega^T\Omega=I}} ||\Omega A-B||_F,
\label{eq:procrustes}
\end{equation}
where for us, $\Omega$ will be a to-be-determined rotation matrix and $\lVert \cdot \rVert_F$ is the Frobenius norm.

Without loss of generality, $\bar{X_0}$ is assumed to be the origin, $(0,0,0)$ and
${X_0}$ is the mean of $\tp$\footnote{In practice, this produces an good initial guess
of translation for \eqref{eq:Pose}}. The translation $p$ is thus given by
\begin{align}
    \bar{X_0} &= R{X_0} + p , \ \bar{X_0}=[0,0,0]^\transpose \nonumber\\
    p &= -R{X_0}.
\end{align}
The rest of the problem can be formulated as:
\begin{align}
    \left\|H_T^L{X_i}-\bar{X_i}\right\|^2_2 &= \left\| \begin{bmatrix} R & p\\ \zeros & 1
        \end{bmatrix} \begin{bmatrix} {X_i} \\ 1 \end{bmatrix} - 
        \begin{bmatrix} \bar{X_i} \\ 1\end{bmatrix}\right\|_2^2 \\
        &= \left\|R{X_i}+p-\bar{X_i}\right\|^2_2 = \left\|R{X_i}^{\prime}-\bar{X_i}\right\|^2_2,\nonumber
\end{align}
\begin{align}
    \sum_{i=1}^4 \left\|H_T^L{X_i}-\bar{X_i}\right\|^2_2 &= \sum_{i=1}^4
    \left\|R{X_i}^{\prime}-\bar{X_i}\right\|^2_2\nonumber\\
                                                     &= \left\|
                                                     (R{X_1}^{\prime}-\bar{X_1})\vdots\
                                                     \cdots \ 
                                                     \vdots(R{X_4}^{\prime}-\bar{X_4})
                                                     \right\|_F^2 \nonumber\\
                                                     &= \left\|R{\X}-\bar{\X} \right\|^2_F,
\end{align}
where ${X_i}^{\prime} = {X_i}-{X_0}$, 
        ${\X}=\begin{bmatrix} {X_1}^\prime& {X_2}^\prime&
            {X_3}^\prime &{X_4}^\prime\end{bmatrix}$ and $\bar{\X}=\begin{bmatrix}
        \bar{X_1}& \bar{X_2}&  \bar{X_3} &\bar{X_4}\end{bmatrix}$.
The problem is then
\begin{align}
    R^\ast = \argmin_{\substack{R:\ R^TR=I}} \left\| R{\X} -\bar{\X}\right\|_F^2.
\end{align}

\noindent By the procrustes optimization problem~\cite{gower1975generalized}, we have
a closed form solution: 
\begin{align}
    M &= \bar{\X}{\X}^\transpose = U \Sigma V^\transpose \\ 
    R^\ast &= U V^\transpose.
\end{align}

\begin{remark}
    To estimate ${\X}$, we project the target point cloud $\tp$ along a principal axis of a
 Principal Components Analysis (PCA)~\cite{price2006principal}. Using the 2D projected point cloud, we use \textit{RANSAC}
    to regress lines to determine target edges and solve for the intersections of the lines, to obtain an initialization of the vertices. The smallest variance of the principal axis is then used for the $\epsilon$ in \eqref{eq:JKHcost}.
    If the number of edge points is less than three, or any of edges fails when
    regressing a line, the cluster is marked as invalid.
\end{remark}

%% file: Decoding.tex
\begin{figure}[t]%
\centering
\subfloat[][]{%
\label{fig:codeword1}%
\includegraphics[width=0.15\columnwidth]{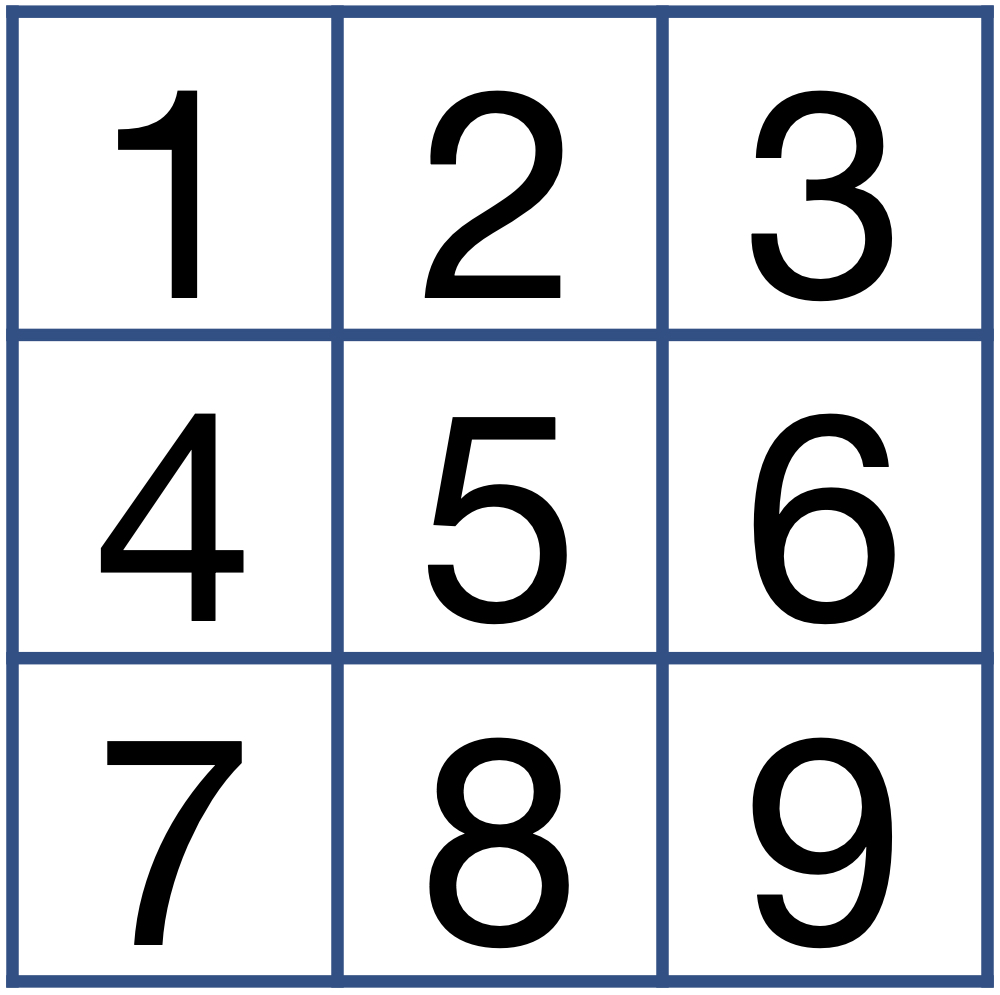}}%
\hspace{12pt}%
\subfloat[][]{%
\label{fig:codeword2}%
\includegraphics[width=0.15\columnwidth]{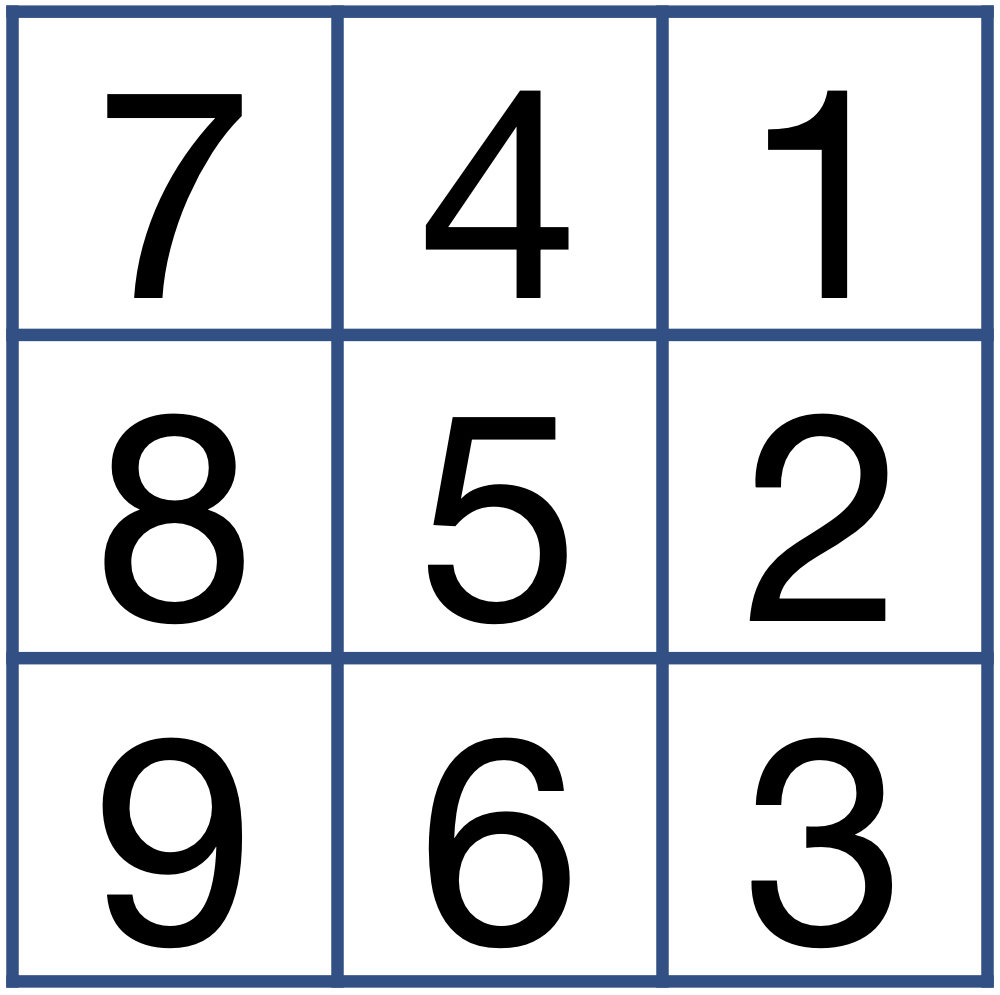}}
\hspace{12pt}%
\subfloat[][]{%
\label{fig:codeword3}%
\includegraphics[width=0.15\columnwidth]{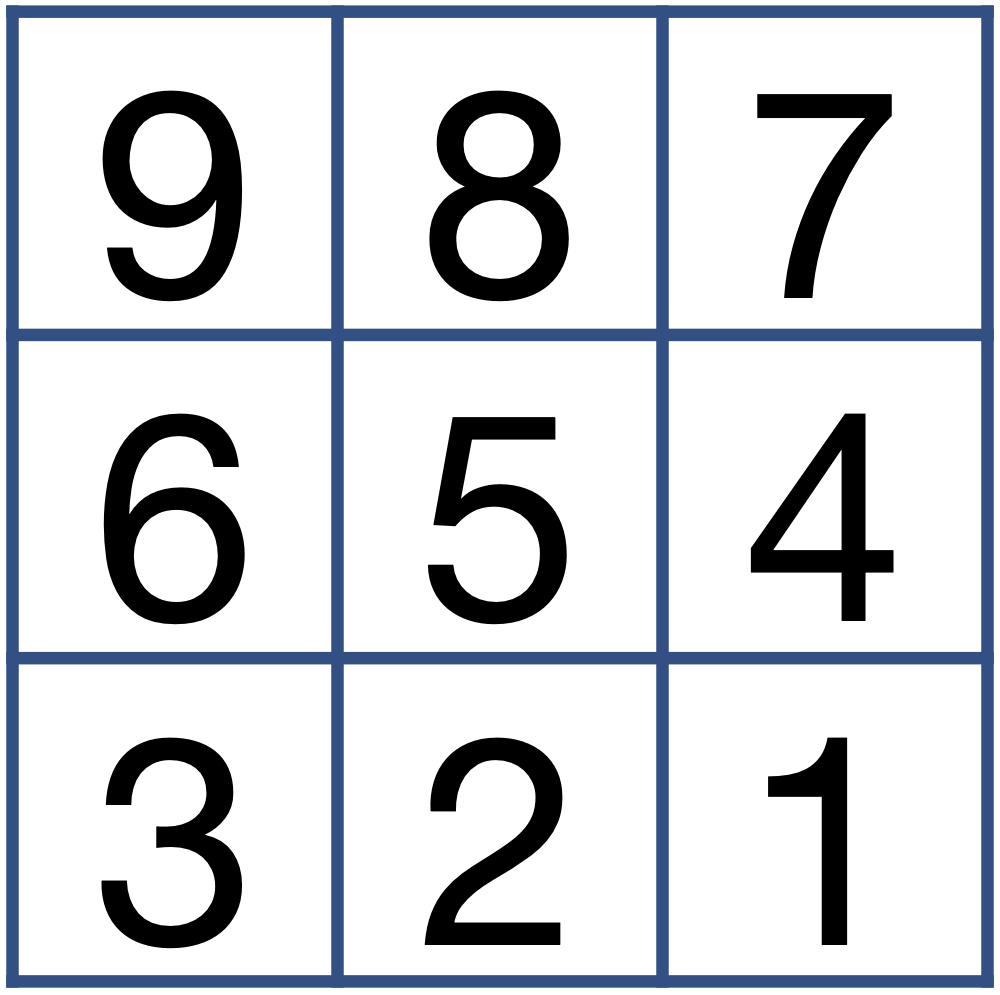}}%
\hspace{12pt}%
\subfloat[][]{%
\label{fig:codeword4}%
\includegraphics[width=0.15\columnwidth]{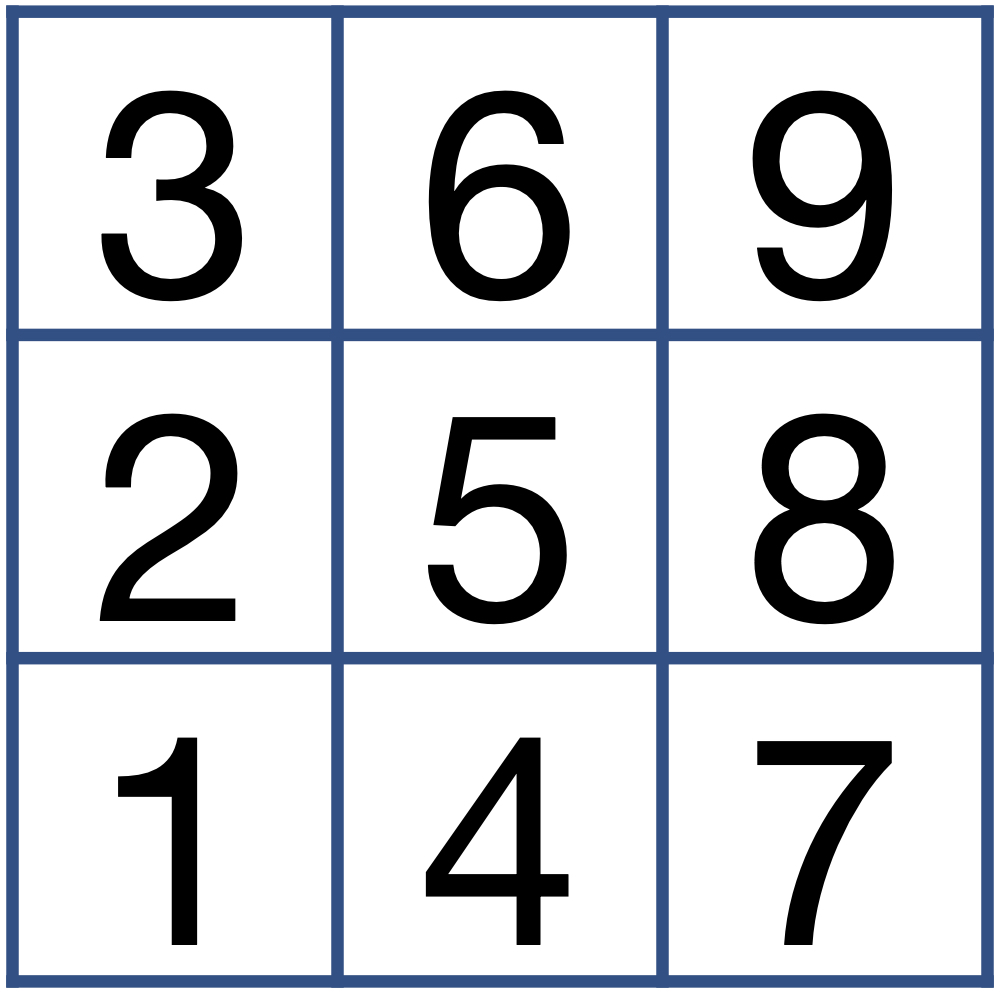}}%
\caption[]{Before decoding, the estimated rotation about the normal axis is only
known modulo $90^\circ$, which means \subref{fig:codeword1} to \subref{fig:codeword4}
yield the same normal vector. Accounting for the three possible rotations of
$(\pm90^\circ, 180^\circ)$, results in 4 possible continuous functions in the
function dictionary. When computing the inner product to the correct id of a
LiDARTag, only one of the four functions is correct. From the correct function, the
modulo $90^\circ$ ambiguity is removed.
}
\label{fig:codeword_rotation}%
\end{figure}

\section{Function Construction and Tag Decoding}
\label{sec:Decoding}
In Sec.~\ref{sec:PoseEstimation}, we defined a template at the \lidar origin,
estimated $H_T^L$, and we thus have the projected point cloud. Specifically, the
projected point cloud is located at the \lidar origin inside the template on the
$y$-$z$ plane with the thickness being the sensor noise on the $x$-axis. Due to the sparsity of the point cloud, we construct a continuous function in an inner product space (RKHS) for the projected point cloud~\cite{MGhaffari-RSS-19}. For each \lidart in the tag family, we pre-compute four
continuous functions to account for four possible rotations, as shown
in Fig.~\ref{fig:codeword_rotation}, consequently, resulting in a function dictionary. Each function is constructed by converting each pixel of the tag image to a point in $\reals^3$, see~\cite{githubLiDARTag} for implementation detail.
Finally, we compute the inner product of the estimated
function and each function in the dictionary. The largest
inner product is the ID of the \lidartN, and the ambiguity of rotation in
Sec.~\ref{sec:PoseEstimation} is removed.

Let 
$\overtilde{\calx} := \{(\overtilde{p_i},\ell(\overtilde{p_i})) | \overtilde{p_i} \in \reals^3~\text{and}~\ell(\overtilde{p_i})\in \Ical \}_{i=1}^{M}$ be a collection of projected points, where $M$ is the number of points. In this work, we use the intensity as our information inner product space as $\Ical = \mathbb{R}$ and the inner product, $\langle \cdot, \cdot \rangle_{\Ical}$, is just the scalar product of intensity values. Therefore, the labels are simply the intensity values. The continuous function of $\overtilde{\calx}$ is defined as
\begin{equation}
    f(\cdot) = \sum_{i=1}^M \ell(\overtilde{p_i})k(\cdot, \overtilde{p_i}),
\end{equation}
where $k:\reals^3\times\reals^3\rightarrow \reals$ is the kernel of an
RKHS~\cite{MGhaffari-RSS-19}. 

Given another continuous function $g$ of point cloud 
$\overtilde{\Zcal} := \{(\overtilde{p_j},\ell(\overtilde{p_j})) | \overtilde{p_j} \in \reals^3~\text{and}~\ell(\overtilde{p_j})\in \Ical \}_{j=1}^{N}$, where $N$ is the number of points. The inner
product of $f$ and $g$ is 
\begin{equation}
    \label{eq:InnerProduct}
    \langle f, g\rangle = \sum_{i=1}^M \sum_{j=1}^N \langle \ell(\overtilde{p_i}),
    \ell(\overtilde{p_j})\rangle_\Ical k(\overtilde{p_i}, \overtilde{p_j}).
\end{equation}
The kernel $k$ is modeled as the squared exponential kernel
~\cite[Chapter 4]{rasmussen2006gaussian}: 
\begin{equation}
\label{eq:kernel}
    k(\overtilde{p_i}, \overtilde{p_j}) = 
    \sigma^2 \text{exp}\left(-\frac{1}{2}(\overtilde{p_i}-\overtilde{p_j})^\transpose 
    \Lambda (\overtilde{p_i}-\overtilde{p_j})\right), 
\end{equation}
where $\sigma^2$ is the signal variance (set to $1e5$) and $\Lambda$ is an isotropic diagonal length-scale matrix with its diagonal entry set to the inverse of squared half of the bit size of a \lidartN: $1/(t/(2(d+4)))^2$. Let $t$ be the \lidart size, and the
$d$-bit tag family is used ($d+4$ bits, including its boundaries). Then the bit size is $t/(d+4)$. 



After applying the kernel trick to~\eqref{eq:InnerProduct}, we get~\cite{clark2020nonparametric}
\begin{align}
\label{eq:newscalar}
    \langle f,g\rangle = \sum_{i=1}^M \sum_{j=1}^N \,  k_{\Ical}(\ell(\overtilde{p_i}),\ell(\overtilde{p_j})) \cdot k(\overtilde{p_i}, \overtilde{p_j}),
\end{align}
where 
\begin{equation}
\label{eq:kernel_intensity}
    k_{\Ical}(\overtilde{p_i}, \overtilde{p_j}) = 
    \text{exp}\left(-\frac{(\ell(\overtilde{p_i})-\ell(\overtilde{p_j}))^2}{2l_{\Ical}^2} \right), 
\end{equation}
and the length-scale $l_{\Ical}$ is set to $10$ ($0 \leq \ell(\overtilde{p_i}) \leq 255$). 

\begin{remark}
To fully utilize the projected point cloud of \lidart returns, we extend the planar \lidart to a 3D \lidart based on the intensity value of each point in the point cloud. 
The linear transformation is defined as:
\begin{equation}
    \overtilde{p_i} = 
    \begin{bmatrix}
    \overtilde{x_i}\\\overtilde{y_i}\\\overtilde{z_i}
    \end{bmatrix} = 
    \begin{bmatrix}
       1&0&0&\frac{t}{2(d+4)I_{max}} \\
       0&1&0&0\\
       0&0&1&0
   \end{bmatrix}
    \begin{bmatrix}
    \bar{x_i}\\\bar{y_i}\\\bar{z_i}\\\bar{i_i}
    \end{bmatrix},
\end{equation}
where $I_{max}$ is the maximum intensity of the point cloud and $t/(d+4)$ is the bit size.
\end{remark}

\begin{remark}
    If the fitting error~\eqref{eq:JKHcost} in Sec.~\ref{sec:PoseEstimation} is
    greater than $10\%$ of the number of points in the cluster or it is not able to
    decode the potential cluster in Sec.~\ref{sec:Decoding}, this cluster is marked
    as invalid.
\end{remark}

\begin{remark}
Reproducing Kernel Hilbert Spaces have been widely used in the Representer
Theorem~\cite{scholkopf2001generalized, wahba1990spline, kimeldorf1971some} for
various regularization problems, such as function estimation, classification and
Support Vector Machines (SVM). These are typically high- or infinite-dimensional
problems that while mathematically feasible, often appear to be not practically
computable. With the help of RKHS and the Representer Theorem, the solutions to these
problems can be formulated in lower-dimensional subspaces spanned by the
``representers'' of the data.
\end{remark}

%% file: Experiments.tex
\section{Experimental Results}
\label{sec:results}
We now present experimental evaluations of the proposed LiDARTag. In this work, we
choose an easel as our 3D object to support the tag. Additionally,
fiducial markers from the tag16h6 family of \atagN3 are used, with sizes of 1.2, 0.8, 0.61 meters, as shown in Fig.
\ref{fig:LiDARTag}. We do not compare the proposed \lidart system with
camera-based tag systems because it is unfair to compare depth estimation from a
\lidar with depth estimation from a monocular camera. All experiments are conducted
with a \velodyne and an Intel RealSense camera rigidly attached to the torso of a
Cassie-series bipedal robot as shown in Fig.~\ref{fig:torso}. We use the Robot
Operating System (ROS)~\cite{quigley2009ros} to communicate and synchronize between
sensors. The \lidart system runs faster than 100 Hz on a laptop equipped with
Intel$^\text{\textregistered}$ Core$^{\text{TM}}$ i7-9750H CPU @ 2.60 GHz, which is
similar to the processor on a robot coming to the market.

Datasets are collected in a cluttered laboratory to evaluate detection performance
and a spacious outdoor facility, M-Air~\cite{MAir}, equipped with a
motion capture system to validate pose estimation and ID decoding. Additionally,
false positives are evaluated on the
Google Cartographer indoor dataset~\cite{hess2016real} and the outdoor Honda H3D
datasets~\cite{360LiDARTracking_ICRA_2019}.

\begin{figure*}[t!]%
\centering
\subfloat[]{%
    \label{fig:MocapImgClose}%
\includegraphics[height=0.2\textwidth]{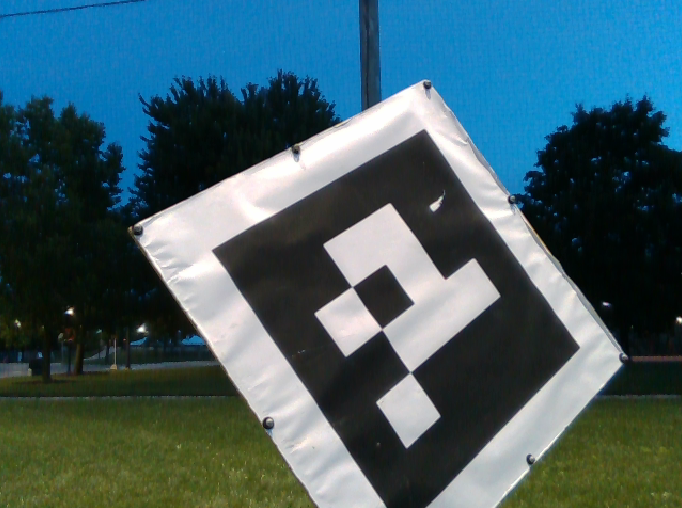}}~
\hspace{3pt}%
\subfloat[]{%
    \label{fig:MocapTagClose}%
\includegraphics[trim=0 0 100 0,clip,height=0.2\textwidth]{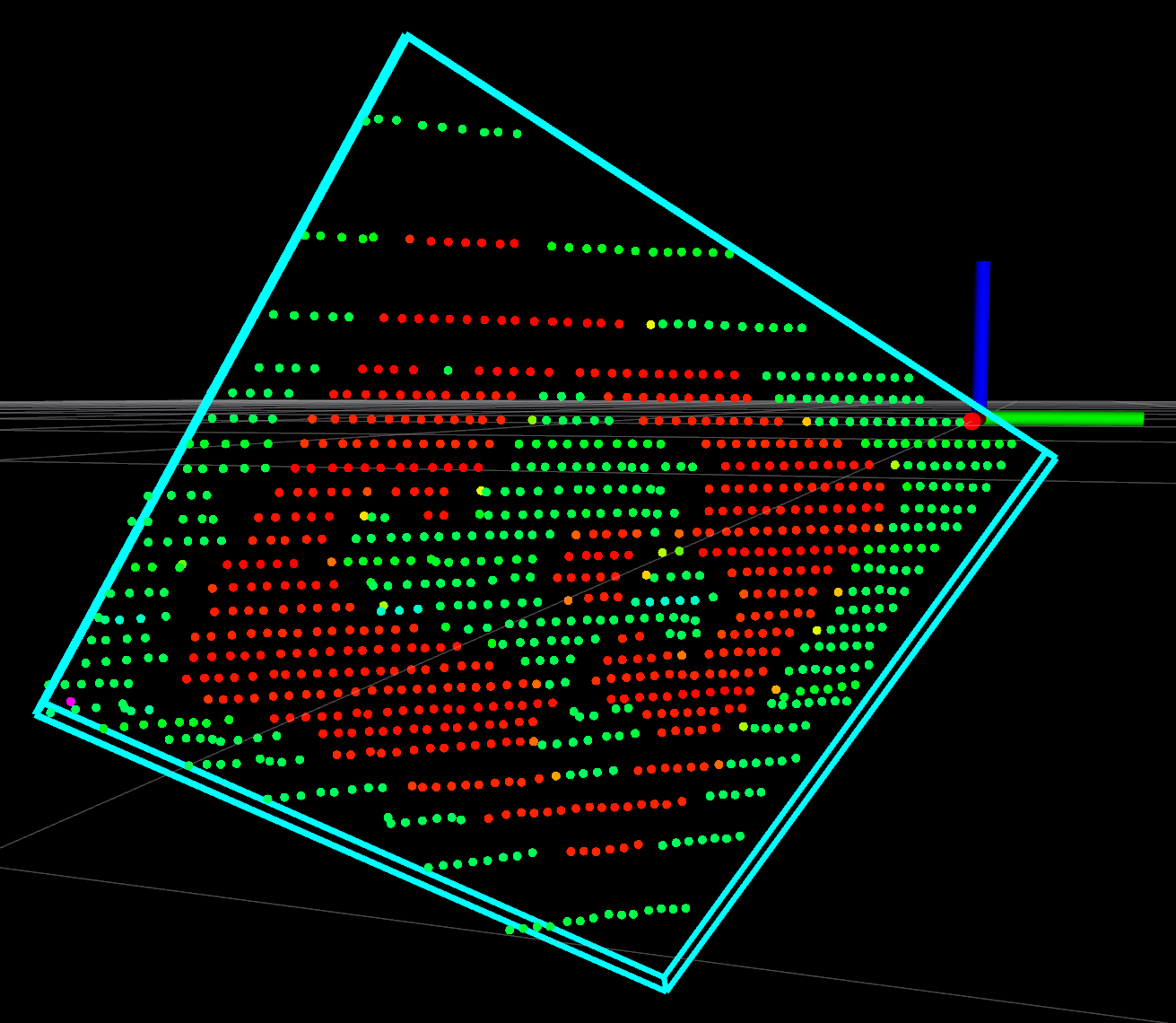}}~
\hspace{3pt}%
\subfloat[]{%
    \label{fig:MocapImgFar}%
\includegraphics[height=0.2\textwidth]{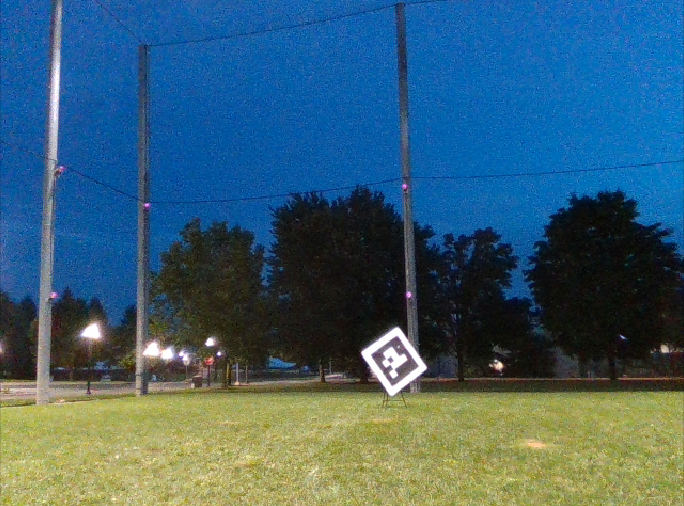}}~
\hspace{3pt}%
\subfloat[]{%
    \label{fig:MocapTagFar}%
\includegraphics[trim=0 0 10 0,clip,height=0.2\textwidth]{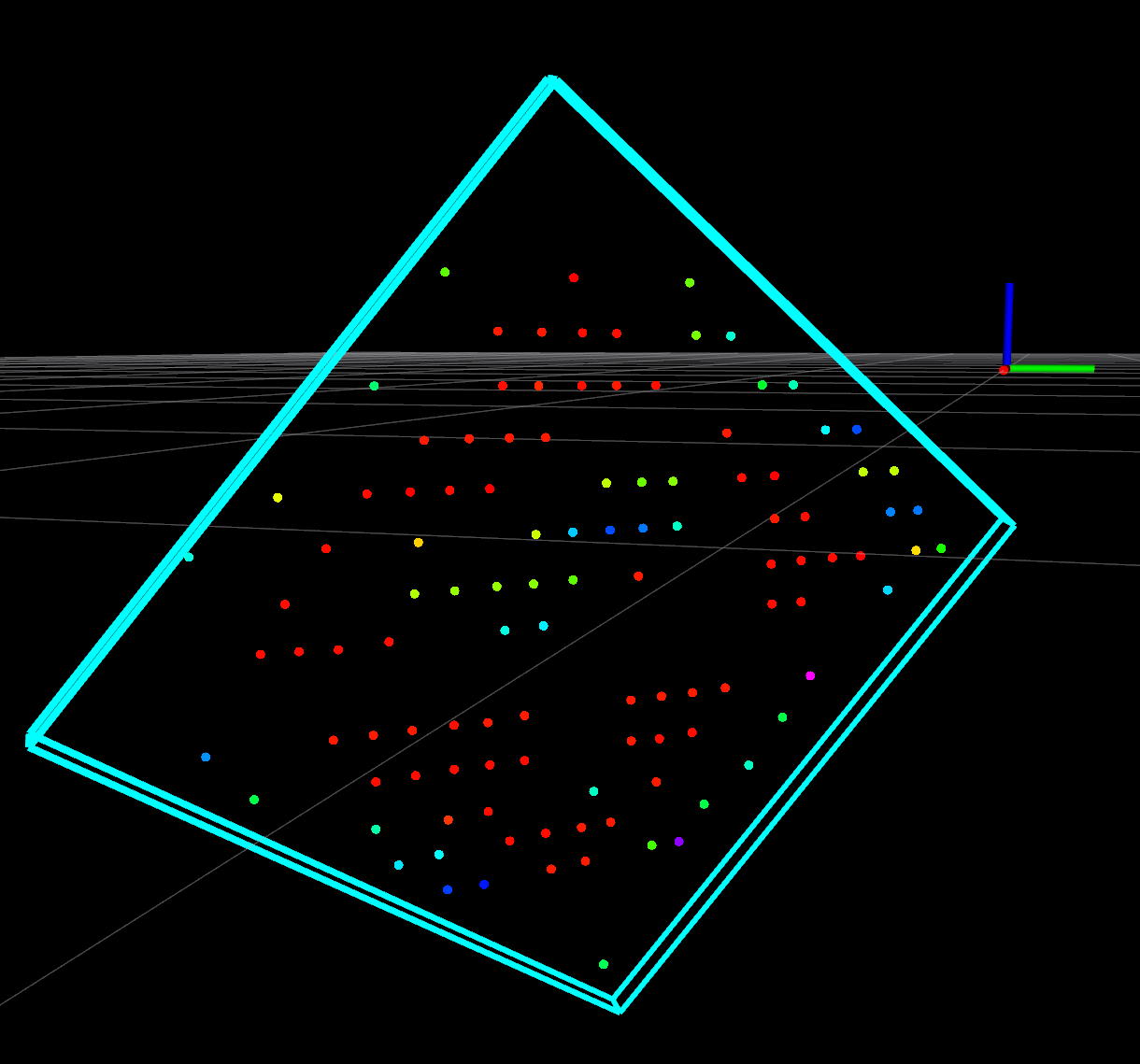}}~%
\caption[]{
    \subref{fig:MocapImgClose} and \subref{fig:MocapImgFar} are images of the tag
    placed at 2 and 14 meters away from a Cassie-series robot.
    \subref{fig:MocapTagClose} and \subref{fig:MocapTagFar} describe the results of
    projecting the template (green box) from the \lidar origin to the tag's returns
    by the poses of \lidart at 2 and 16 meters, respectively. While
    \subref{fig:MocapTagFar} shows much sparser \lidar returns than
    \subref{fig:MocapTagClose} due to the farther distance, we are still able to
    accurately estimate the pose and its ID. Compared to ground truth provided by 30
    motion capture cameras, the resulting poses are a few millimeters off in
    translation and a few degrees off in rotation.
}%
 \label{ExpFigures}%
\end{figure*}

\input{TablePose}
\subsection{Pose Evaluation and Decoding Accuracy}
\label{sec:PoseEvalutation}
A motion capture system developed by Qualisys is used as a proxy for ground truth
poses. The setup consists of 30 motion capture cameras with markers attached to tags,
a LiDAR and a camera, as shown in Fig.~\ref{fig:torso}. Datasets are
collected at various distances and angles. Each of the datasets contains
images (20 Hz) and scans of point clouds (10
Hz). The 1.2-meter target is placed at distances from 2 to 14 meters in 2 meter
increments. At each distance, data is collected with a target face-on to the \lidar and
another dataset with the target rotated by 45 degrees. More results and videos are on GitHub, see \cite{githubLiDARTag}.

The optimization problem in \eqref{eq:Pose} is solved with the
method of moving asymptotes (MMA) algorithm~\cite{svanberg2002class,
svanberg1987method} provided in NLopt library~\cite{NLopt}. We use the optimized
\lidart pose to project the template at the \lidar origin onto the \lidartN's returns
to show the qualitative results of pose estimation. Figure~\ref{fig:MocapTagClose}
and Figure~\ref{fig:MocapTagFar} show the pose of a tag at 2 meter and 16 meter,
respectively. Even though the farther marker has much sparser \lidar returns, by
lifting the returns to an RKHS space, we are capable of correctly identifying its ID. In particular, a 1.2-meter target placed at 16 meters and rotated by 45 degrees is the detection limit of our \velodyneN. However, for a \lidar with a different number of beams or points, the detection limit is subject to change. The more beams or points, the greater is the detection range. Using our \lidarN, the detectable angle is a little smaller than camera-based marker due to the sparse point clouds.
Table~\ref{tab:PoseAccuracy} compares quantitatively the  pose estimation between the
proposed \lidart and ground truth. The translation error is reported in millimeters, and rotation error $\xi$ is represented as geodesic distance
\footnote{The shortest path between two points on the $\SO(3)$ group.} 
in degrees~\cite{huynh2009metrics}:
\begin{equation}
\label{eq:so3err}
    \xi = \|\Log(\bar{R} \overtilde{R}^\transpose)\|,
\end{equation}
where $\lVert \cdot \rVert$ is the Euclidean norm, $\bar{R}$ and $\overtilde{R}$ are the ground truth and estimated rotation matrices, respectively, and $\Log(\cdot)$ is the  logarithm map in the Lie group $\SO(3)$.

\subsection{\lidart System and Speed Analysis}
\label{sec:SpeedAnalysis}
The computation time and cluster analysis of each step of the pipeline is shown in
Table~\ref{tab:timing}. Indoors, we have fewer clusters because detected features
are closer to each other resulting in many of them being clustered together. The computation time in an outdoor environment is faster than indoors because more clusters are rejected in the early stage due to the sparsity of the clusters, see Table ~\ref{tab:remainingcClusters}. In both environments, the system achieves
real-time performance (at least 100 Hz). 
\input{TableTiming}
\input{TableClusterRemoval}

\begin{figure*}[t!]%
\centering
\subfloat[]{%
    \label{fig:TwoTagsLabImg}%
\includegraphics[height=0.175\textwidth]{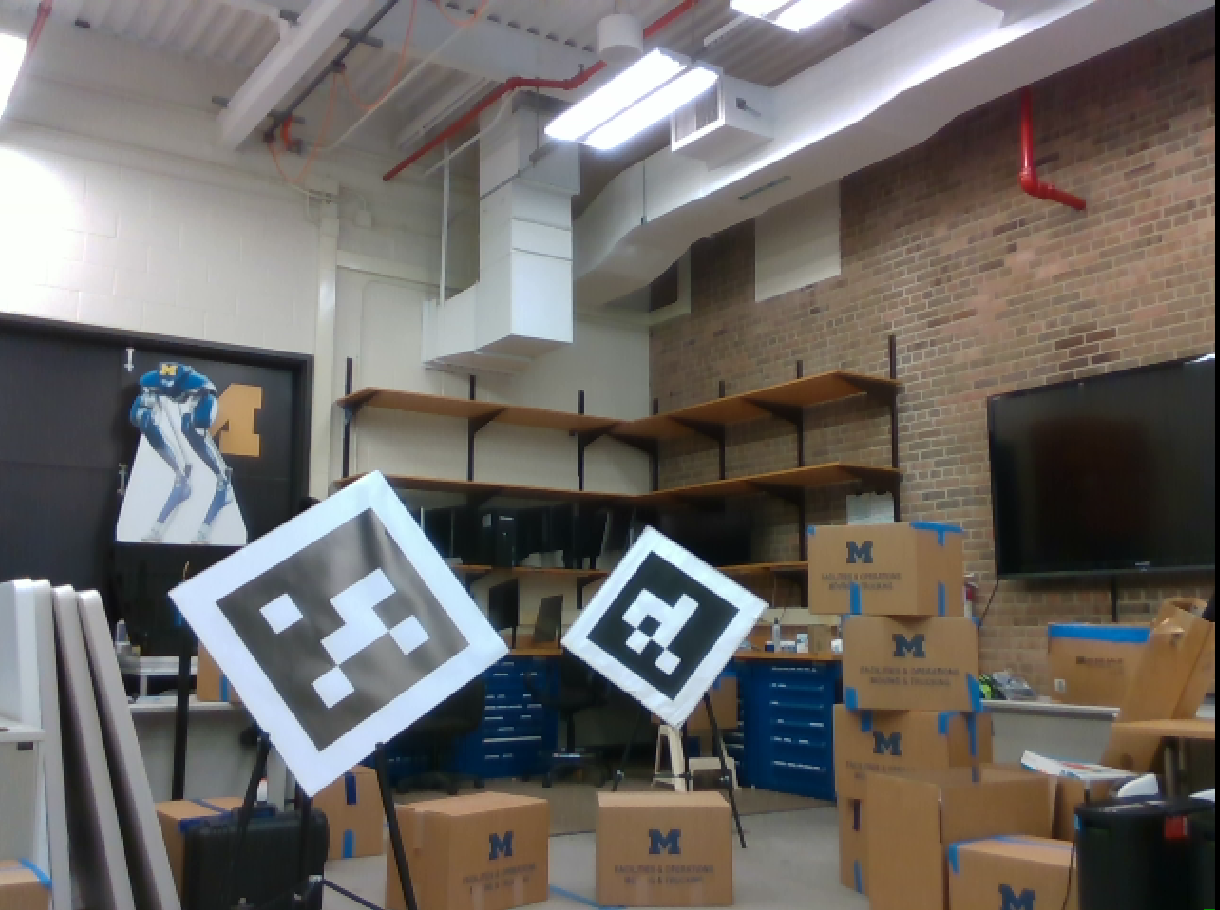}}
\hspace{3pt}%
\subfloat[]{%
    \label{fig:TwoTagsInLabLiDAR}%
\includegraphics[height=0.175\textwidth]{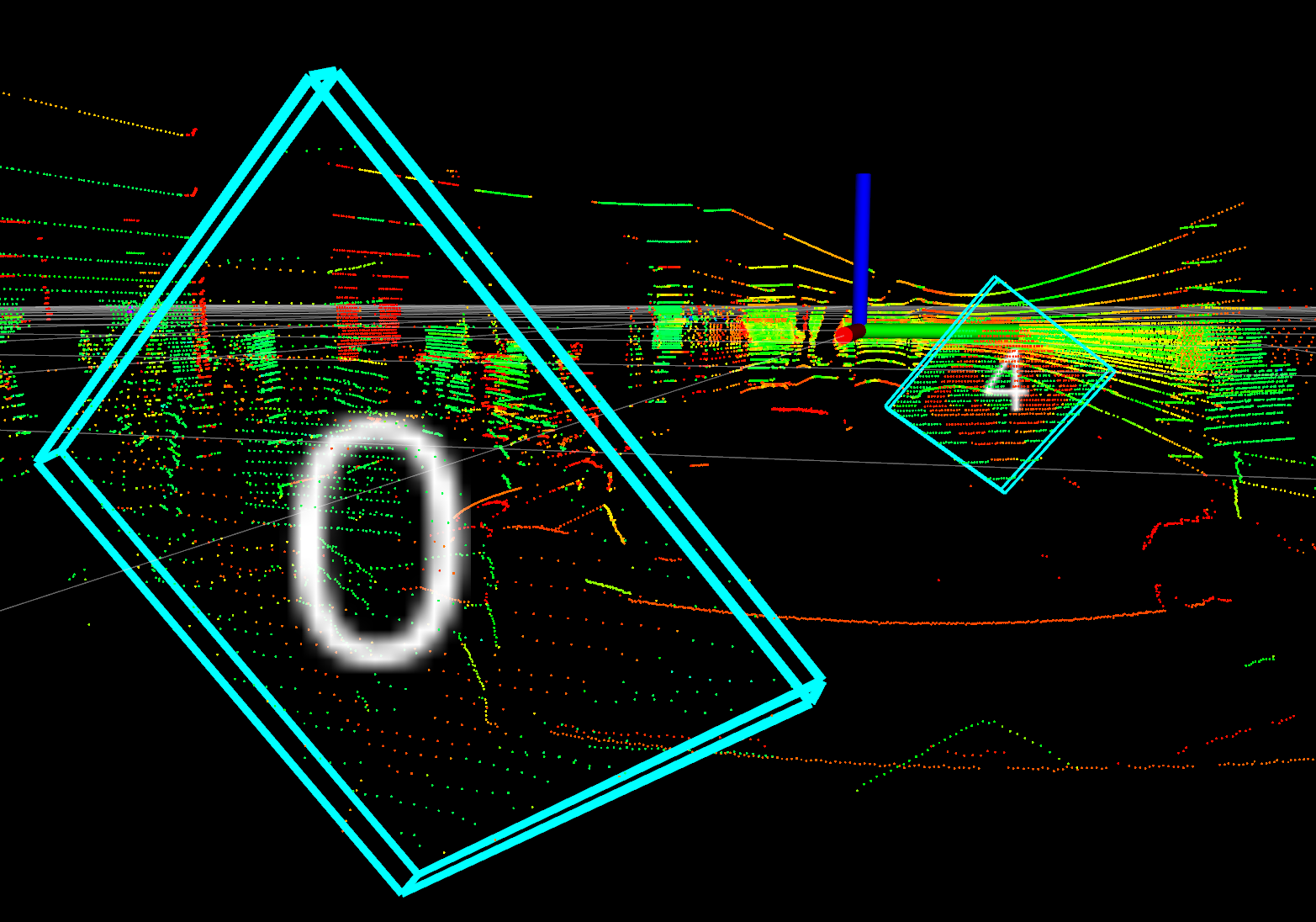}}
\hspace{3pt}%
\subfloat[]{%
    \label{fig:TwoTagsOutsideImg}%
\includegraphics[trim=20 0 10 0,clip,height=0.175\textwidth]{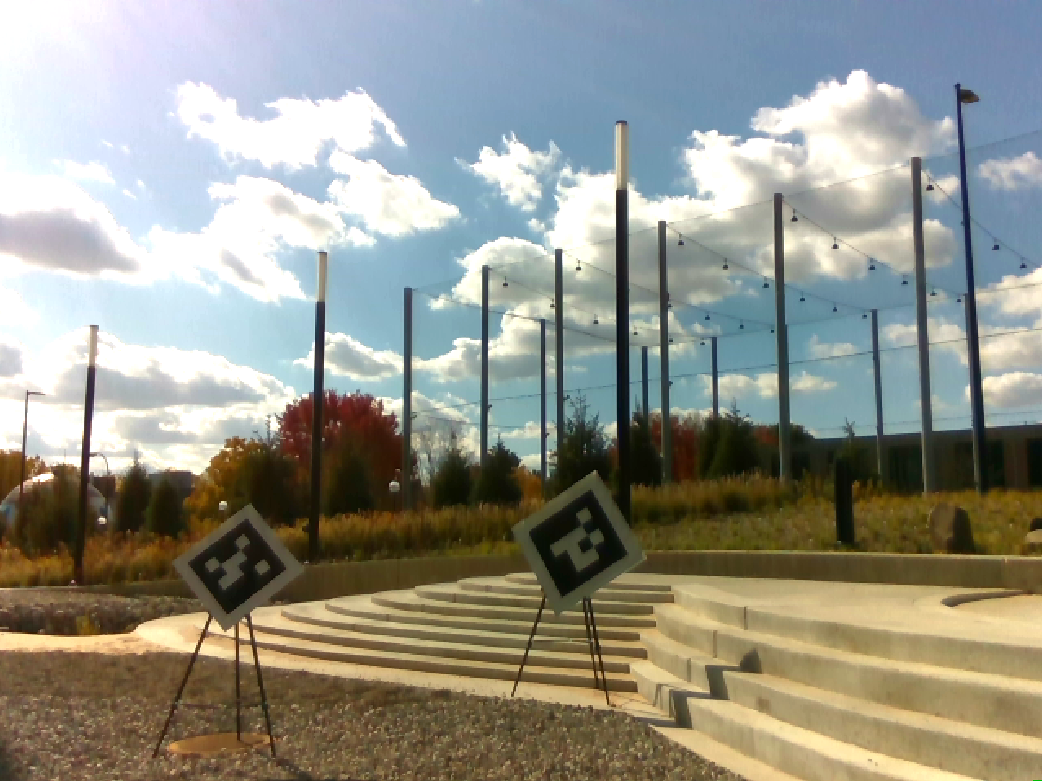}}
\hspace{3pt}%
\subfloat[]{%
    \label{fig:TwoTagsOutsideLiDAR}%
\includegraphics[trim=30 0 100 0,clip,height=0.175\textwidth]{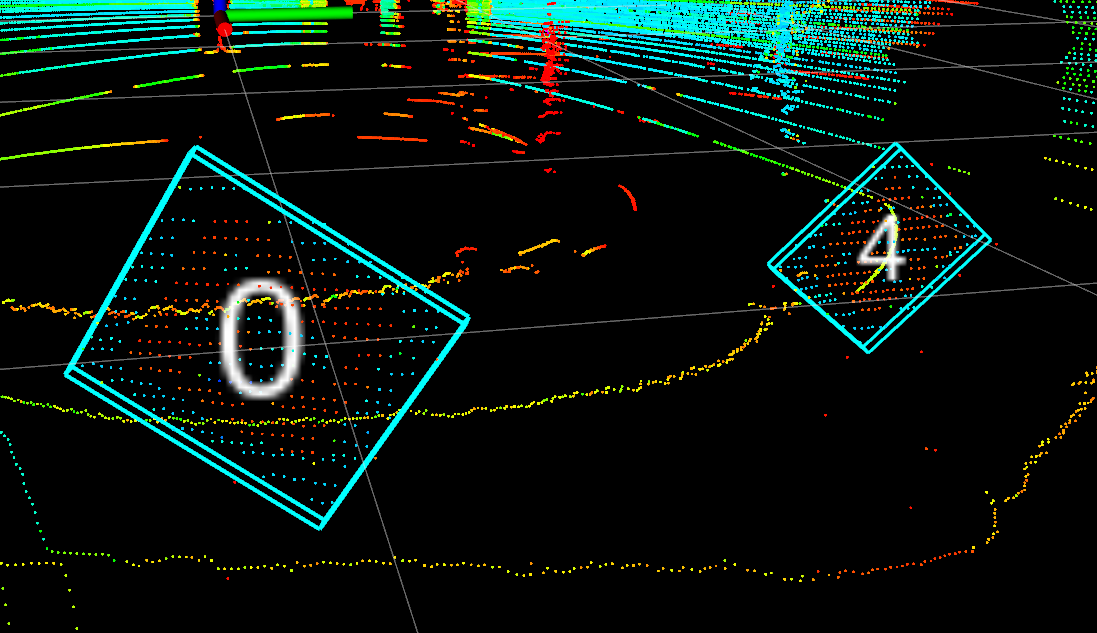}}%
\caption[]{
    \subref{fig:TwoTagsLabImg} and \subref{fig:TwoTagsOutsideImg} image a 0.8
        and a 0.6 meter tag placed in a cluttered indoor laboratory and a spacious
        outdoor environment. \subref{fig:TwoTagsInLabLiDAR} and
        \subref{fig:TwoTagsOutsideLiDAR} show the algorithm successfully detects the
        two markers of different sizes indicated by cyan boxes.
}%
\label{fig:TwoTags}%
\end{figure*}

The original double sum in \eqref{eq:InnerProduct} takes over 140 milliseconds for
each decoding process. To speed up the process, the inner sum of the double sum is
transformed to a matrix and then to a vector form. For
more details, see our implementation~\cite{githubLiDARTag}. These two modifications boost the speed to 8.5 ms. However,
this is still not fast enough because for each remaining cluster,
\eqref{eq:InnerProduct} needs to be computed with all the tags in the function
dictionary. Threading Building Blocks library (TBB)~\cite{TBB} is therefore used to
further speed up this process to 2.4 ms. All together, the whole
process was sped up by a factor of 60, from 144 ms to 2.4 ms. Furthermore, we also investigated the speed
performance of employing a k-d tree data structure~\cite{blanco2014nanoflann}. A summary
Table~\ref{tab:decodingMethods} is presented, showing that use of a k-d tree did not improve performance. 
\input{TableDoubleSum}

\subsection{False Positive Analysis}
We have chosen some public datasets containing no ARTag features (i.e., no \lidartsN) so that there cannot be any false negatives and any detection of a \lidartN in the datasets is consequently a false positive. To better verify the proposed \lidart algorithm, cluttered indoor scenes and
crowded outdoor scenes are both necessary. The Google Cartographer
indoor dataset~\cite{hess2016real} and Honda H3D
outdoor dataset~\cite{360LiDARTracking_ICRA_2019} were therefore used to validate the
false positive rate of the proposed system. The Cartographer was collected
with two Velodyne VLP-16 \lidars in the Deutsches Museum. We took the longest three
sequences consisting of more than 350 thousand \lidar scans. Each scan contains about
30,000 points. We used the same algorithm with the same parameters and the full function dictionary that was used to detect the two tags in Fig.~\ref{fig:TwoTags} (true positives were detected), and no false positives (i.e., no \lidartsN) were detected in the three sequences.
\input{TableFPs}

The Honda H3D dataset was collected by a 64-beam Velodyne \lidar and consists of 160
crowded and highly interactive traffic scenes in the San Francisco Bay Area. We evaluated
on all sequences, resulting in 29 thousand \lidar scans. Each scan consists of more
than 130 thousand points. Zero targets were extracted by the detector. The results
are shown in Table~\ref{tab:falsePositives}. Additionally, false positives removed by
each step are provided in Table~\ref{tab:remainingcClusters}. Last but not the least,
Fig.~\ref{fig:TwoTags} shows that the detector is able to detect markers of
different sizes both in a cluttered indoor scene and a spacious outdoor scene.

%% file: TablePose.tex

\squeezeup
\begin{table*}[b]
\center
\caption{
Decoding accuracy of the RKHS method and pose accuracy of the fitting method. The ground truth is provided by a motion capture
system with 30 motion capture cameras. The distance is in meters. The translation error is in
millimeters and rotation error is the misalignment angle, \eqref{eq:so3err}, in degrees.
}
\label{tab:PoseAccuracy}
\resizebox{2\columnwidth}{!}{%
    \begin{tabular}{|c|c|c|c|c|c|c|c|c|c|}
\hline
\multicolumn{5}{|c|}{\textbf{Face-on to LiDAR}}    & \multicolumn{5}{c|}{\textbf{Rotated at 45 degrees}} \\ \hline
Distance         & No. Scans & No. Wrong ID   & Translation Error & Rotation Error & Distance         & No. Scans & No. Wrong ID   & Translation Error & Rotation Error \\ \hline
2.15            & 73    & 0        & 14.03 & 0.44  & 2.13             & 74    & 0        & 0.27  & 0.05  \\ \hline
4.29            & 72    & 0        & 10.13 & 0.67  & 3.95             & 134   & 0        & 0.52  & 0.34  \\ \hline
5.90            & 81    & 0        & 16.23 & 0.44  & 5.93             & 137   & 0        & 0.36  & 0.05  \\ \hline
7.97            & 78    & 0        & 1.32  & 0.21  & 7.92             & 126   & 0        & 0.26  & 0.32  \\ \hline
10.12           & 87    & 0        & 1.64  & 0.40  & 10.38            & 130   & 0        & 4.91  & 1.03  \\ \hline
12.14           & 69    & 0        & 2.07  & 0.36  & 12.12            & 71    & 0        & 5.78  & 0.39  \\ \hline
13.87           & 35    & 1        & 2.81  & 10.48 & 14.08            & 49    & 2        & 1.98  & 15.92 \\ \hline
\textbf{Summary} & No. Scans & Wrong ID Ratio & Translation Error & Rotation Error & \textbf{Summary} & No. Scans & Wrong ID Ratio & Translation Error & Rotation Error \\ \hline
\textbf{mean}   & 70    & 0.202 \% & 6.891 & 2.149 & \textbf{mean}    & 103   & 0.276 \% & 1.744 & 2.586 \\ \hline
\textbf{std}    & 16.88 & --       & 6.418 & 4.577 & \textbf{std}     & 37.25 & --       & 2.076 & 5.888 \\ \hline
\textbf{median} & 73    & --       & 2.81  & 0.44  & \textbf{median}  & 126   & --       & 0.52  & 0.34  \\ \hline
\end{tabular}
}
\end{table*}

%% file: TableTiming.tex
\begin{table}[]
\center
\caption{
    This table averages all the datasets we collected and
    describes computation time of each step for indoors and outdoors.
}
\label{tab:timing}
\resizebox{1\columnwidth}{!}{%
\begin{tabular}{|c|c|c|c|}
\hline
\multicolumn{4}{|c|}{\textbf{Outdoor}}                                     \\ \hline
No. Points & No. Features          & No. Clusters      & Total Computation \\ \hline
51717        & 2179          & 271            & 114.86 Hz         \\ \hline
PoI Clustering  & Fill In Clusters & Point Check       & Plane Fitting     \\ \hline
2.63 ms         & 0.3 ms           & 0.00 ms           & 0.27  ms          \\ \hline
Line Fitting    & PCA              & Pose Optimization & Tag Decoding      \\ \hline
0.01 ms         & 0.03 ms          & 0.48 ms           & 3.34 ms           \\ \hline
\multicolumn{4}{|c|}{\textbf{Indoor}}                                      \\ \hline
No. Point Cloud & No. Features          & No. Clusters      & Total Computation \\ \hline
54277        & 1820          & 225            & 102.41 Hz         \\ \hline
PoI Clustering  & Fill In Clusters & Point Check       & Plane Fitting     \\ \hline
3.28 ms         & 0.22 ms          & 0.00 ms           & 0.15 ms           \\ \hline
Line Fitting    & PCA              & Pose Optimization & Tag Decoding      \\ \hline
0.01 ms         & 0.01 ms          & 0.42 ms           & 2.47 ms           \\ \hline
\end{tabular}
}
\end{table}

%% file: TableClusterRemoval.tex
\begin{table}[]
\center
\caption{
    This table takes into account all the data we collected and shows numbers of
    rejected clusters in each step in different scenes. Additionally, we also report
false positive rejection for Cartographer and H3D dataset.
}
\label{tab:remainingcClusters}
\resizebox{0.9\columnwidth}{!}{%
    \begin{tabular}{|c|c|c|}
\hline
\multicolumn{3}{|c|}{\textbf{Outdoor}}                           \\ \hline
No. Min. Return     & No. Max. Return     & No. Plane Fitting    \\ \hline
247.72              & 3.41                & 14.71                \\ \hline
No. Boundary Points & No. Pose Estimation & No. Decoding Failure \\ \hline
4.10                & 0.48                & 0.00                 \\ \hline
\multicolumn{3}{|c|}{\textbf{Indoor}}                            \\ \hline
No. Min. Return     & No. Max. Return     & No. Plane Fitting    \\ \hline
76.44               & 1.12                & 0.00                 \\ \hline
No. Boundary Points & No. Pose Estimation & No. Decoding Failure \\ \hline
8.14                & 1.80                & 1.16                 \\ \hline
\multicolumn{3}{|c|}{\textbf{Indoor Cartographer Dataset}}       \\ \hline
No. Min. Return     & No. Max. Return     & No. Plane Fitting    \\ \hline
65.76               & 0                   & 1.90                 \\ \hline
No. Boundary Points & No. Pose Estimation & No. Decoding Failure \\ \hline
0.35                & 0                   & 0                    \\ \hline
\multicolumn{3}{|c|}{\textbf{Outdoor H3D Dataset}}               \\ \hline
No. Min. Return     & No. Max. Return     & No. Plane Fitting    \\ \hline
713.35              & 8.72                & 44.41                \\ \hline
No. Boundary Points & No. Pose Estimation & No. Decoding Failure \\ \hline
2.74                & 0.38                & 0                    \\ \hline
\end{tabular}
}
\end{table}

%% file: TableDoubleSum.tex
\begin{table}[t]
\center
\caption{
    The original double sum in \eqref{eq:InnerProduct} is too slow to achieve a
    real-time application. This table compares different methods to compute the
    double sum, in which the TBB stands for Threading Building Blocks library from
Intel. Additionally, we also apply a k-d tree data structure to speed up the querying
process; the k-d tree, however, does not produce fast enough results. The unit in
the table is milliseconds.
}
\label{tab:decodingMethods}
\resizebox{0.9\columnwidth}{!}{%
\begin{tabular}{|c|c|c|}
\hline
Original Double Sum & Matrix Form     & Vector From \\ \hline
144.18              & 67.11           & 8.51        \\ \hline
TBB Original Form   & TBB Vector Form & TBB k-d tree \\ \hline
35.68               & 2.40            & 5.73        \\ \hline
\end{tabular}
}
\end{table}

%% file: TableFPs.tex
\begin{table}[t]
\center
\caption{
    This table shows the numbers of false positive rejection of the proposed algorithm.
    We validated the rejection rate on the indoor Google Cartographer dataset and
    the outdoor Honda H3D datasets. The former has two VLP-16 Velodyne \lidar and the
latter has one 64-beam Velodyne \lidarN.
}
\label{tab:falsePositives}
\resizebox{0.95\columnwidth}{!}{%
\begin{tabular}{|c|c|c|}
\hline
                    & Google Cartographer & Honda H3D             \\ \hline
Scene               & Indoor Museum       & Crowed Driving Scenes \\ \hline
Duration            & 150 minutes         & 48 minutes               \\ \hline
No. Scans           & 350 thousand       & 29 thousand             \\ \hline
No. False Positives & 0                   & 0                     \\ \hline
\end{tabular}
}
\squeezeup
\end{table}

%% file: Conclusion.tex
\section{Conclusion and Future Work}
\label{sec:conclusion}
We presented a novel and flexible fiducial marker system specifically for point
clouds. The developed fiducial tag system runs in real-time (faster than 100 Hz)
while it can handle a full scan of raw point cloud from the employed \velodyne (up to
120,000 points per scan). Each step of the proposed system was extensively analyzed
and evaluated in both cluttered indoor as well as spacious outdoor environments.
Furthermore, the system can be operated in a completely dark environment.

The \lidart pose estimation block deploys an $L_1$-inspired cost function. It achieved
millimeter accuracy in translation and a few degrees of error in rotation compared
to ground truth data collected by a motion capture system with 30 motion capture
cameras. The sparse \lidar returns on a \lidart are lifted to a continuous function
in a reproducing kernel Hilbert space where the inner product is used to determine
the marker's ID, and this method achieved 99.7$\%$ accuracy. The rejection of
false positives was evaluated on the Google Cartographer indoor dataset and the Honda
H3D outdoor dataset. No false positives were detected in over 379 thousand \lidar
scans.

The presented fiducial marker system can also be used with cameras and has been
successfully used for \lc calibration in~\cite{huang2019improvements}
and~\cite{githubFileExtrinsic}. Additionally, the system is able to detect various
marker sizes, whereas camera-based fiducial markers support one marker size at a
time. In the future, we shall use the developed \lidart within SLAM systems to
provide robot state estimation and loop closures. Because of different inherent
properties of \lidars and cameras, it would also be interesting to fuse a
camera-based tag system and the proposed \lidart system. Currently, the proposed algorithm assumes the point cloud has been motion compensated; how to adopt motion distortion into the algorithm is an interesting direction for future work. Furthermore, if a
dataset has been collected and labeled, a deep-learning architecture can replace
the process of \lidart detection, thus offering another interesting area for future
research.

